%% file: main.tex
\documentclass[10pt,twocolumn,letterpaper]{article}

\usepackage[pagenumbers]{cvpr}        %

\usepackage{iftex}
\ifPDFTeX\else
  \usepackage[OT1]{fontenc}
\fi

\input{preamble}
\AtBeginDocument{\crefname{table}{Table}{Tables}}  %
\makeatletter
\renewcommand\paragraph{\@startsection{paragraph}{4}{\z@}{1.6ex \@plus .5ex \@minus .2ex}{-1em}{\normalfont\normalsize\bfseries}}
\g@addto@macro\normalsize{%
  \setlength{\abovedisplayskip}{3pt plus 2pt minus 1pt}\setlength{\belowdisplayskip}{3pt plus 2pt minus 1pt}%
  \setlength{\abovedisplayshortskip}{2pt plus 2pt}\setlength{\belowdisplayshortskip}{4pt plus 2pt minus 2pt}}
\makeatother
\input{lang_en}

\definecolor{cvprblue}{rgb}{0.21,0.49,0.74}
\usepackage[pagebackref,breaklinks,colorlinks,allcolors=cvprblue]{hyperref}

\def\confName{CVPR}
\def\confYear{2027}

\title{X-3DComposer: Feed-Forward 3D Scene Composition\\ with Any Assets from Any Views}

\author{Zidian Qiu \qquad Ancong Wu\thanks{Corresponding author.}\\[2pt]
Sun Yat-sen University\\[1pt]
{\small qiuzd@mail2.sysu.edu.cn, \ wuanc@mail.sysu.edu.cn}\\[3pt]
{\small Code: \href{https://github.com/qiuzidian/X-3DComposer}{X-3DComposer}}}

\begin{document}
\maketitle
\input{sec/0_abstract}
\input{sec/1_intro}
\input{sec/2_related}
\input{sec/3_method}
\input{sec/4_experiments}
\input{sec/5_conclusion}
{
    \small
    \bibliographystyle{ieeenat_fullname}
    \bibliography{main}
}

\input{sec/X_suppl}

\end{document}

%% file: preamble.tex
\usepackage{multirow}
\usepackage{bm}
\usepackage{colortbl}
\definecolor{oursgray}{gray}{0.93}
\definecolor{ourstint}{RGB}{229,239,250}
\definecolor{blockband}{gray}{0.92}
\newlength{\tabgroupgap}
\newcommand{\oursrow}{\rowcolor{ourstint}}
\newcommand{\blockrow}[2]{\rowcolor{blockband}\multicolumn{#1}{c}{#2}}  %
\newcommand{\best}[1]{\textbf{#1}}
\newcommand{\second}[1]{\underline{#1}}

\newcommand{\bt}{\mathbf{t}}
\newcommand{\bs}{\mathbf{s}}
\newcommand{\bp}{\mathbf{p}}
\newcommand{\bq}{\mathbf{q}}
\newcommand{\bx}{\mathbf{x}}
\newcommand{\bo}{\mathbf{o}}   %
\newcommand{\be}{\mathbf{e}}
\newcommand{\bz}{\mathbf{z}}   %
\newcommand{\bl}{\boldsymbol{\ell}}
\newcommand{\LR}{L^{R}}
\newcommand{\Lo}{L^{\bo}}
\newcommand{\Phat}{\hat{P}}
\newcommand{\Rhat}{\hat{P}^{R}}   %
\newcommand{\ohat}{\hat{P}^{\bo}}  %
\newcommand{\diag}{\operatorname{diag}}
\newcommand{\vggto}{VGGT-$\Omega$}

\newcommand{\fitbox}[1]{\resizebox{\ifdim\width>\linewidth\linewidth\else\width\fi}{!}{#1}}

%% file: lang_en.tex
\newcommand{\Lmeshes}{Meshes}
\newcommand{\Llayout}{Layout}
\newcommand{\Lmethod}{Method}
\newcommand{\Linput}{Cond.}  %
\newcommand{\Lnative}{Native}
\newcommand{\Lscenemaker}{SceneMaker$^\dagger$}
\newcommand{\Lours}{Ours}

\newcommand{\Lnone}{none}

\newcommand{\Lsingle}{Single-view}
\newcommand{\Lmulti}{Multi-view}
\newcommand{\Lmultisource}{IndoorMix}

\newcommand{\Lvariant}{Variant}
\newcommand{\Lfull}{Full model}
\newcommand{\Lnocrops}{w/o object crops}
\newcommand{\Lnoid}{w/o identity embeddings}
\newcommand{\Lnophys}{w/o physics constraints}
\newcommand{\Lnoground}{w/o grounding head}
\newcommand{\Lnopre}{w/o pretrained weights}
\newcommand{\Ltime}{Time (s)}

\newcommand{\Lsinglepano}{One panorama}
\newcommand{\Lmultipano}{Two panoramas}

\newcommand{\Ldataset}{Dataset}
\newcommand{\Lscenes}{\#Scenes}
\newcommand{\Lreconscans}{InternScenes}

\newcommand{\Lphysics}{Physics}
\newcommand{\Lpen}{Pen.$\downarrow$}  %
\newcommand{\Lmem}{Mem.\ (GB)}

\newcommand{\LsecEst}{Estimated conditions}
\newcommand{\LsecGT}{Ground-truth conditions}
\newcommand{\LsecSV}{Single-view input}
\newcommand{\LsecMV}{Multi-view input}
\newcommand{\Lsinglelc}{single-view}
\newcommand{\Lmultilc}{multi-view}

%% file: sec/0_abstract.tex
\begin{abstract}
Composing an editable 3D scene from images requires an asset for each object and a layout that places the assets. 3D generation and CAD retrieval now supply the assets, but layout still lacks a general solution: scene generation methods couple it with generation, and pose estimators handle one object at a time and need true-size CAD models, depth, or iterative refinement. We present X-3DComposer, a feed-forward model that predicts the layout separately from asset generation. Given one or more images of a scene, the object masks, and an asset for each object from any source, it jointly predicts the rotation, translation, and per-axis scale of all objects in a single forward pass of about one second. The key idea is to treat objects as views: renders of the assets are processed together with the scene images by a multi-view geometry model adapted from \vggto{}, and each object pose follows in closed form from the cameras predicted for its renders. Camera intrinsics, camera poses, and depth are optional inputs. When the meshes produced by five scene generation methods are kept fixed and only their layout is replaced, the layouts of X-3DComposer are more accurate than the native layouts of these methods. Compared with the stand-alone layout model SceneMaker, it reduces the scene-level Chamfer distance by 52\% on IndoorMix, our test set from six indoor datasets, and by 28\% on 3D-FUTURE.
\end{abstract}

%% file: sec/1_intro.tex
\section{Introduction}
\label{sec:intro}

Composing an editable 3D scene of separate objects from one or more images is needed in content creation, embodied AI, and robot manipulation. Such a scene consists of a 3D asset for each object and the pose and scale of each asset. Assets are increasingly easy to obtain. 3D generators~\cite{xiang2025trellis,sam3dteam2025sam3d} produce high-quality meshes from images, and CAD retrieval can provide exact or similar CAD models. This paper studies the other half of the problem, scene layout: given the assets, estimate the pose and scale of each one so that the composed scene matches the images.\looseness=-1

\begin{figure}[t]
  \centering
  \includegraphics[width=\linewidth]{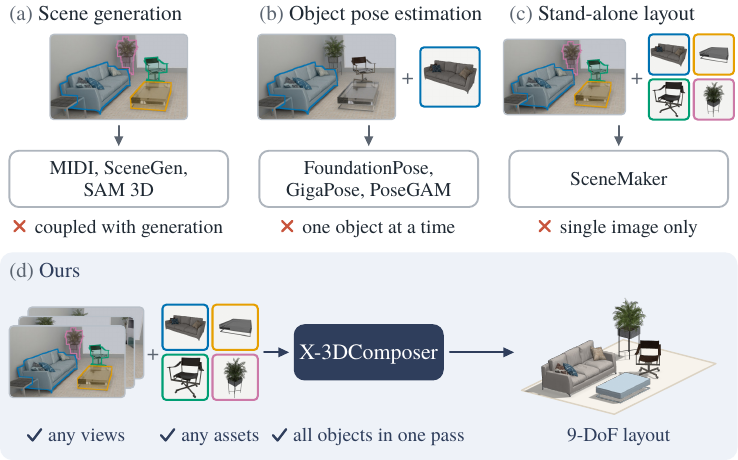}
  \caption{Approaches to scene layout: (a) scene generation, (b) object pose estimation, (c) stand-alone layout models, and (d) X-3DComposer, which places all objects in a single forward pass, from any number of views and with assets from any source.}
  \label{fig:teaser}
\end{figure}

Existing methods fall into three groups (\cref{fig:teaser}). The first group consists of scene generation methods~\cite{huang2025midi,meng2026scenegen,sam3dteam2025sam3d,li2026mvsam3d,xia2026fire3d}, which output a layout together with the object geometry. This layout is a by-product of generation and cannot be applied to assets from other sources. The second group takes a CAD model as given. Object pose estimators~\cite{wen2024foundationpose,labbe2022megapose} handle one object at a time, assume a true-size CAD model, often require depth, and refine each pose by render-and-compare. CAD alignment methods~\cite{gumeli2022roca} also estimate scale, but they too treat each detected object independently. The third group consists of stand-alone layout models~\cite{shi2025scenemaker,liao2026spoon,zhang2026easy3dfy}, which separate layout from generation but rely on iterative sampling, dense correspondences, or per-scene optimization; most accept only a single image.\looseness=-1

A general layout method should accept assets from any source and place all objects jointly, in a single forward pass, from one or more views. None of the methods above meets all of these requirements. Multi-view geometry models~\cite{wang2025vggt,wang2026vggtomega} already do this for cameras: they predict the cameras of any number of views in a common frame at once. We observe that layout can be cast as the same task by treating objects as views. We render each asset from a few viewpoints with known cameras in its own frame and feed the renders to a multi-view geometry model together with the scene images. The model predicts a camera for every view in a common scene frame. Each render camera is then known in the asset frame and predicted in the scene frame, and the object pose follows from the two in closed form. All objects are thus placed jointly, without sampling or optimization, and since the model needs only renders of an asset, it applies to any asset. Predicting a camera for each view is also the pretraining task of these models, which lets us start from the pretrained \vggto~\cite{wang2026vggtomega}, keep its architecture, and add few parameters.

Whole scenes raise further difficulties. The model must know which object each render belongs to, even when objects look identical, so each object is assigned an identity embedding that links its masks, its crops from the scene images, and its renders. Generated or retrieved assets often differ in proportions from the real objects, and we therefore predict one scale per axis. The input also varies from one image to several, with or without camera intrinsics, camera poses, and depth. We treat these as optional geometric conditions. Our contributions are as follows:
\begin{itemize}
  \item We present X-3DComposer, a feed-forward model that predicts the layout of all objects (rotation, translation, and per-axis scale) in a single forward pass of about one second, given any number of scene images and assets from any source.
  \item We extend camera-based pose prediction~\cite{chen2026posegam,qiu2026pano3dcomposer} from single objects to whole scenes: the crops and renders of all objects are fed as extra views to a multi-view geometry model adapted from \vggto{}, with identity embeddings and optional geometric conditions, and the object poses are read out in closed form from the predicted cameras.
  \item With the meshes fixed and only the layout changed, our layouts are more accurate than the native layouts of five scene generation methods, and our scene-level Chamfer distance is 52\% lower than that of SceneMaker~\cite{shi2025scenemaker} on our IndoorMix test set and 28\% lower on 3D-FUTURE.
\end{itemize}

%% file: sec/2_related.tex
\section{Related Work}
\label{sec:related}

\paragraph{Compositional 3D scene generation and reconstruction.}
One line of work generates the geometry of several objects together with their layout. MIDI~\cite{huang2025midi} and SceneGen~\cite{meng2026scenegen} generate all objects of a scene jointly, while SAM 3D~\cite{sam3dteam2025sam3d} and its multi-view extension MV-SAM3D~\cite{li2026mvsam3d} generate a mesh for each object and predict its rotation, translation, and scale. FIRE3D~\cite{xia2026fire3d} also outputs the objects together with their layout. Such a layout applies only to the assets generated with it. Our model does not depend on a particular generator. Another line of work first obtains the assets and then solves for the layout. CAST~\cite{yao2025cast} places each generated object in the scene point cloud with an alignment model and refines the poses by optimization under physics constraints. Given a single image, SceneMaker~\cite{shi2025scenemaker} samples the object poses iteratively from a pose model conditioned on object geometry. Mira-Scene~\cite{sun2026mirascene} and Easy-3Dfy~\cite{zhang2026easy3dfy} solve for the transformation from dense correspondences between the object and an estimated point cloud, and Lucida~\cite{qin2026lucida} adjusts the poses iteratively with a vision-language model. SPOON~\cite{liao2026spoon} handles uncalibrated multi-view input by guiding the pose sampling of SAM 3D with a \vggto{} reconstruction, followed by per-scene optimization. Pano3DComposer~\cite{qiu2026pano3dcomposer} feeds object crops and asset renders to VGGT and predicts a transformation with per-axis scale for each object separately, from panoramic input. These methods handle one object at a time or rely on iterative procedures, whereas our model places all objects in a single forward pass.\looseness=-1

\begin{figure*}[t]
  \centering
  \includegraphics[width=1.0\linewidth]{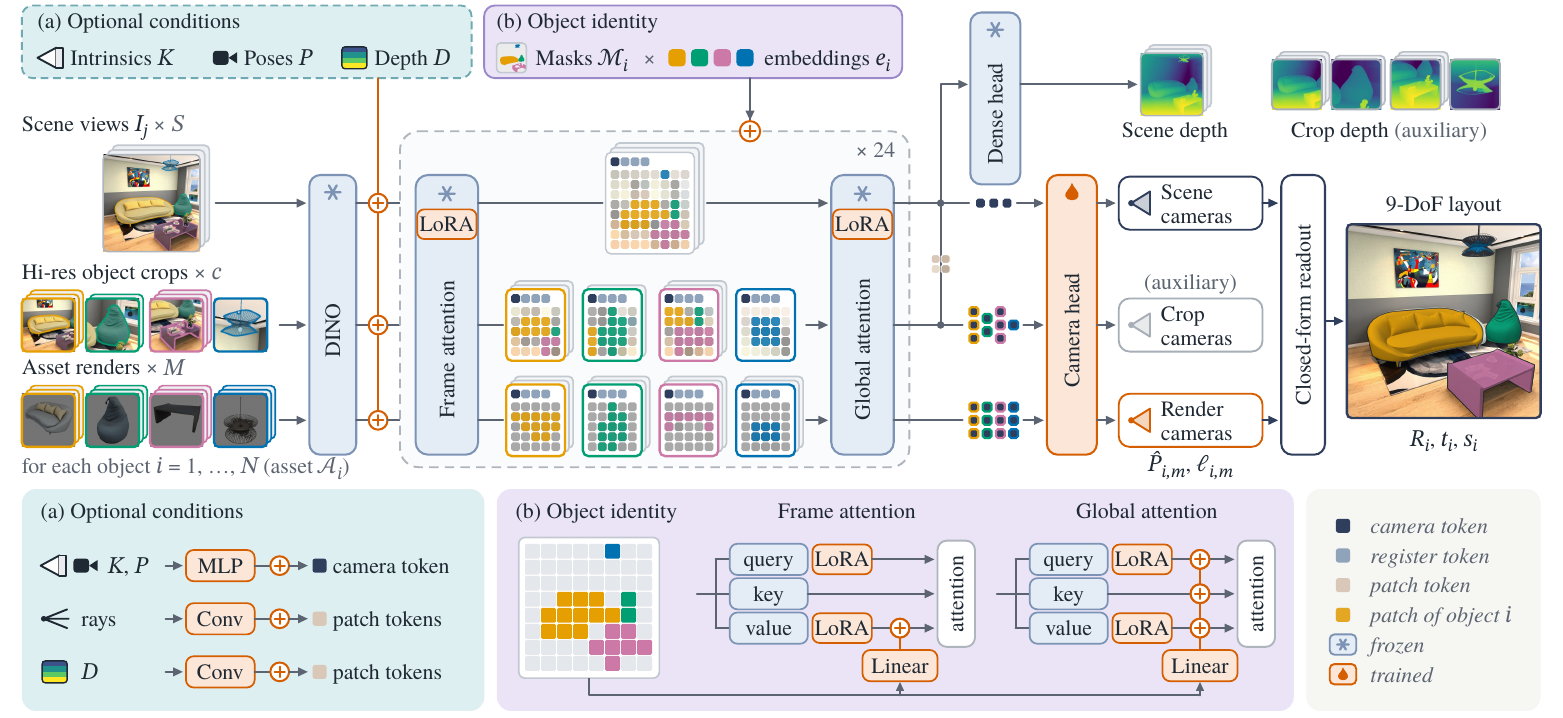}
  \caption{Overview. Scene views, object crops, and asset renders are processed as one set of views by a model adapted from \vggto{}. (a) The optional conditions and their encoders. (b) Object identity: the embeddings $\be_i$, added where the masks $\mathcal{M}_i$ lie, link the masks, crops, and renders of each object. The layout $(R_i,\bt_i,\bs_i)$ is read out in closed form from the predicted render cameras $\Phat_{i,m}$ and log-scales $\bl_{i,m}$.}
  \label{fig:overview}
\end{figure*}

\paragraph{CAD alignment and object pose estimation.}
Single-image CAD alignment methods~\cite{kuo2020mask2cad,gumeli2022roca,gao2023diffcad} retrieve a CAD model for each detected object and align it to the image, and recent methods~\cite{arsomngern2025zeroshotcad,ejaz2026sufleca} solve for pose and scale from feature correspondences between the image and the CAD model. Methods for 6-DoF pose estimation of novel objects assume a true-size CAD model. MegaPose~\cite{labbe2022megapose} and FoundationPose~\cite{wen2024foundationpose} use render-and-compare, GigaPose~\cite{nguyen2024gigapose} uses templates and feature correspondences, and AlignPose~\cite{mikestikova2026alignpose} refines poses over multiple views. Closest to our approach is PoseGAM~\cite{chen2026posegam}. Initialized from VGGT, it treats renders of the CAD model as views with known cameras, regresses the camera of the query image, and derives the 6-DoF object pose from it. In contrast, we place all objects jointly and predict per-axis scales, and thus need no true-size CAD models.

\paragraph{Multi-view geometry models.}
DUSt3R~\cite{wang2024dust3r} and MASt3R~\cite{leroy2024mast3r} regress point maps directly from image pairs, and VGGT~\cite{wang2025vggt} predicts the cameras, depth, and point maps of multiple images in a single forward pass. Later models such as $\pi^3$~\cite{wang2025pi3}, Depth Anything 3~\cite{lin2025depthanything3}, and \vggto~\cite{wang2026vggtomega} follow the same paradigm. Pow3R~\cite{jang2025pow3r}, MapAnything~\cite{keetha2026mapanything}, and OmniVGGT~\cite{peng2025omnivggt} additionally accept camera intrinsics, camera poses, or depth as optional inputs. These models reconstruct the scene as a whole and have no notion of individual objects; we keep their task and optional conditions and add objects as extra views.

%% file: sec/3_method.tex
\section{Method}
\label{sec:method}

\Cref{fig:overview} gives an overview of X-3DComposer; details are in the supplementary material.

\subsection{Problem Definition}
\label{sec:problem}

The required input consists of $S \ge 1$ images of a scene, the \emph{scene views} $\{I_j\}_{j=1}^{S}$; the masks $\mathcal{M}_i$ of the $N$ target objects in these views; and one 3D asset $\mathcal{A}_i$ per object. An asset is a mesh, usually textured, from a CAD library, a retrieval system, or a 3D generator. It need not match its object exactly; the same category and a similar shape suffice. From these inputs, our method predicts the layout of all objects,
\begin{equation}
  \{(R_i,\bt_i,\bs_i)\}_{i=1}^{N} = \mathcal{F}\big(\{I_j\},\,\{\mathcal{M}_i\},\,\{\mathcal{A}_i\};\,K,P,D\big),
  \label{eq:task}
\end{equation}
where $K$, $P$, and $D$ are optional geometric conditions described below. For each object $i$, the layout specifies a rotation $R_i\in SO(3)$, a translation $\bt_i\in\mathbb{R}^3$, and a per-axis scale $\bs_i\in\mathbb{R}^3_{>0}$, nine degrees of freedom in total. A point $\bp$ on the asset is mapped to the scene point
\begin{equation}
  \bx = R_i\,\diag(\bs_i)\,\bp + \bt_i .
  \label{eq:placement}
\end{equation}

\paragraph{Frames and geometric conditions.}
Each asset is first normalized: it is translated so that its bounding box is centered at the origin and scaled uniformly so that the longest side of the box is 1. This defines its \emph{asset frame}. The output poses are expressed in the \emph{scene frame}, the camera frame of the first scene view (\cref{fig:readout}). The scene views may come with three optional geometric conditions: intrinsics $K$, the camera pose $P_j$ of each view relative to the first one (collectively $P$), and depth $D$. They may be measured, for example by an RGB-D sensor or SLAM, or estimated by other models. A camera pose is the rigid transformation from the camera frame to a reference frame; the superscripts $R$ and $\bo$, as in $P_j^{R}$ and $P_j^{\bo}$, denote the rotation and camera center of a pose. If the input fixes a unit of length through depth or camera poses, the output is in that unit; otherwise it is in \emph{normalized units}, in which the mean distance from the scene points to the first camera is 1.

\subsection{Objects as Views}
\label{sec:oav}

\paragraph{Background: multi-view geometry models.}
\vggto~\cite{wang2026vggtomega}, the successor to VGGT~\cite{wang2025vggt}, predicts a camera and a depth map for every input image in a single forward pass, with all cameras expressed in the camera frame of the first image. Each image is encoded into patch tokens by a DINO encoder, and one camera token and several register tokens are prepended. The tokens of all images pass through alternating frame-attention and global-attention blocks. A camera head then regresses the camera of each image from its camera token, and a dense head predicts depth.

\paragraph{Representing objects as views.}
We do not feed the asset geometry to the model. Instead, we render each asset and treat the renders as extra views. The pretrained model already takes images as input, so its ability to relate views carries over to relating an asset to the scene, and no 3D encoder has to be trained from scratch.

\paragraph{Three types of views.}
The model takes three types of views as input (\cref{fig:overview}):
\begin{itemize}
  \item \emph{Scene views}: the $S$ scene images.
  \item \emph{Object crops}: $c$ square regions around each object, cut from the full-resolution scene images. Scene views are downsampled before entering the model, and the crops restore the detail that small objects lose.
  \item \emph{Asset renders}: $M$ images of each asset rendered in its asset frame. The camera pose of the $m$-th render of object $i$, $L_{i,m}=[\LR_{i,m}\mid \Lo_{i,m}]$, and its intrinsics are known by construction.
\end{itemize}
Views of all types are processed alike, and the camera head predicts a camera for each. The masks are not fed to the model as images; they determine where the identity embeddings of the objects are added (\cref{sec:network}).

\begin{figure}[t]
  \centering
  \includegraphics[width=\linewidth]{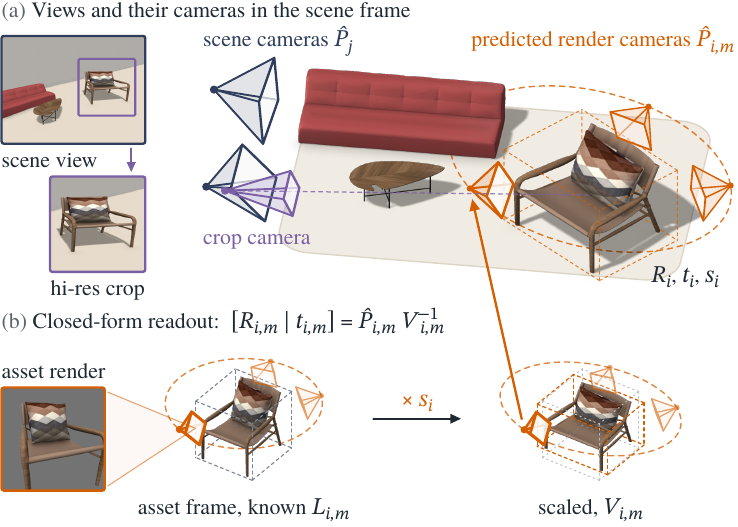}
  \caption{Pose readout. (a) All cameras are predicted in the scene frame. (b) A render camera is $L_{i,m}$ in the asset frame, $V_{i,m}$ after the asset is scaled by $\bs_i$, and $\Phat_{i,m}$ in the scene frame; the transformation between the last two is the object pose.}
  \label{fig:readout}
\end{figure}

\paragraph{Pose readout.}
The \emph{readout} converts the predicted render cameras into object poses. Let $\bq=\diag(\bs_i)\,\bp$ be the coordinates in the \emph{scaled asset frame}, where \cref{eq:placement} becomes the rigid transformation $\bx=R_i\bq+\bt_i$. In this frame, the camera of the $m$-th render is
\begin{equation}
  V_{i,m} = \big[\,\LR_{i,m} \;\big|\; \diag(\bs_i)\,\Lo_{i,m}\,\big],
  \label{eq:vcam}
\end{equation}
that is, its orientation is unchanged and its center is scaled together with the asset (\cref{fig:readout}b). For an isotropic scale this is exact; for an anisotropic scale it is a convention, which the training labels and the readout both follow. The model predicts the camera of each render in the scene frame, $\Phat_{i,m}=[\Rhat_{i,m}\mid\ohat_{i,m}]$, expressed relative to the predicted camera of the first scene view. Since $V_{i,m}$ and $\Phat_{i,m}$ describe the same camera in the scaled asset frame and in the scene frame, each render gives a rigid pose of the object in closed form:
\begin{equation}
  [\,R_{i,m}\mid \bt_{i,m}\,] \;=\; \Phat_{i,m}\,V_{i,m}^{-1},
  \label{eq:readout}
\end{equation}
or, explicitly, $R_{i,m}=\Rhat_{i,m}(\LR_{i,m})^{\top}$ and $\bt_{i,m}=\ohat_{i,m}-R_{i,m}\,\diag(\bs_i)\,\Lo_{i,m}$.

\paragraph{Fusion over renders.}
We average the $M$ pose candidates $(R_{i,m},\bt_{i,m})$ given by \cref{eq:readout} into the fused pose: the translation $\bt_i$ is the mean of the candidate translations, and the rotation $R_i$ is the mean of the candidate rotation matrices, projected back onto $SO(3)$ by SVD. The scale is computed before the poses, since the candidates depend on it. The model outputs a per-axis log-scale $\bl_{i,m}$ for each render (\cref{sec:network}), and the object scale is $\bs_i=\exp\big(\frac{1}{M}\sum_{m}\bl_{i,m}\big)$, used for all render cameras of object $i$ in \cref{eq:vcam}. If the input fixes a unit of length, we estimate one global factor that converts normalized units to this unit and multiply all $\bt_i$ and $\bs_i$ by it. The readout involves no test-time optimization and does not use the crops.

\begin{table*}[t]
  \centering
  \footnotesize
  \setlength{\tabcolsep}{6.6pt}\setlength{\tabgroupgap}{21.8pt}\renewcommand{\arraystretch}{1.08}
  \caption{Layout prediction for generated assets. Each group shares its meshes. Native: the method's own layout. $\dagger$: pose model only.}
  \label{tab:genasset}
  \fitbox{\input{tables/tab_genasset}}
\end{table*}

\subsection{Network}
\label{sec:network}

\paragraph{Backbone and trainable parameters.}
We build on \vggto{} and keep its architecture. The trainable parameters (\cref{fig:overview}) are (i) the output branch of the camera head, (ii) LoRA~\cite{hu2022lora} on the query and value projections of the frame- and global-attention blocks and of the attention blocks in the camera head, and (iii) the condition encoders and identity embeddings introduced below; all other weights are frozen. The output layers of the new modules are zero-initialized, so training starts from the output of the pretrained backbone. The camera head outputs a 9-dimensional camera encoding per view. We add three dimensions for the per-axis log-scale $\bl$, used only for the renders.

\paragraph{Identity embeddings.}
Identical objects cannot be told apart by appearance, so each crop and render must be tied to its object explicitly. We keep a set of learnable identity embeddings and assign one of them, $\be_i$, at random to each object of a scene. An embedding thus only distinguishes the objects within a scene and carries no other meaning. In every view, the identity vector of patch $u$ is the sum of the embeddings of the objects whose masks overlap the patch, weighted by the overlap:
\begin{equation}
  \bz_u = \sum_{i=1}^{N} \alpha_{i,u}\,\be_i ,
  \label{eq:identity}
\end{equation}
where $\alpha_{i,u}\in[0,1]$ is the fraction of the patch covered by the mask of object $i$, so that $\bz_u$ is zero on the background. In scene views and crops, the masks are those given as input; in a render, the mask is the foreground of the rendered asset. After a linear projection, $\bz_u$ is added to the values of every attention block and, in the global-attention blocks, also to the queries and keys.\looseness=-1

\paragraph{Condition encoders.}
For a view with intrinsics and a camera pose, the camera parameters are encoded by an MLP and added to the camera token, and the Pl\"ucker ray of each pixel is embedded by a convolution and added to the patch tokens. With intrinsics only, the rays are computed in the view's own camera frame. A depth map is embedded by a second convolution and added to the patch tokens. The render cameras $L_{i,m}$ are always known and are encoded in the same way, in the asset frame.

\subsection{Training}
\label{sec:training}

Training uses the ground-truth cameras and depth of the scene, as well as the ground-truth pose and scale $(R_i^*,\bt_i^*,\bs_i^*)$ of every object. As in the backbone, all targets are expressed in the normalized units of \cref{sec:problem}.

\paragraph{Supervision of cameras, scales, and poses.}
Scene views keep the losses of the backbone. Renders use the same camera loss, with the label $P^*_{i,m}=[R_i^*\mid\bt_i^*]\,V^*_{i,m}$, where $V^*_{i,m}$ is built from the ground-truth scale as in \cref{eq:vcam}. Each log-scale $\bl_{i,m}$ is supervised with an L1 loss against $\log\bs_i^*$. Crops are supervised with auxiliary camera and depth losses.

The camera loss treats each render separately and does not directly constrain the pose that is finally read out. We therefore also compute the pose candidates differentiably with \cref{eq:readout} and compare two poses of object $i$ by
\begin{equation}
  d_i(R,\bt;R',\bt') \;=\; \frac{\lVert\bt-\bt'\rVert_2}{\lVert\bs_i^*\rVert_2} \;+\; \frac{\lVert R-R'\rVert_F}{\sqrt{2}} .
  \label{eq:pose_err}
\end{equation}
The translation term is divided by the object size $\lVert\bs_i^*\rVert_2$ and is thus independent of the unit of length. We penalize $d_i$ between every candidate and the ground truth, between the fused pose and the ground truth, and between every candidate and the fused pose.

\paragraph{Symmetry, grounding, and physics.}
Objects that are symmetric about the vertical axis, such as round tables and vases, have several equally correct poses. We detect such symmetries for each asset offline and train with the equivalent label closest to the current prediction. A grounding head, used only in training, takes the camera token of each crop and render and predicts which scene-view patches its object occupies. This encourages the model to link crops and renders to their objects in the scene. We also apply two physics constraints to the layout that is read out, penalizing penetration between objects and the offset of floor-supported objects from the floor.

\begin{figure*}[t]
  \centering
  \includegraphics[width=1.0\linewidth]{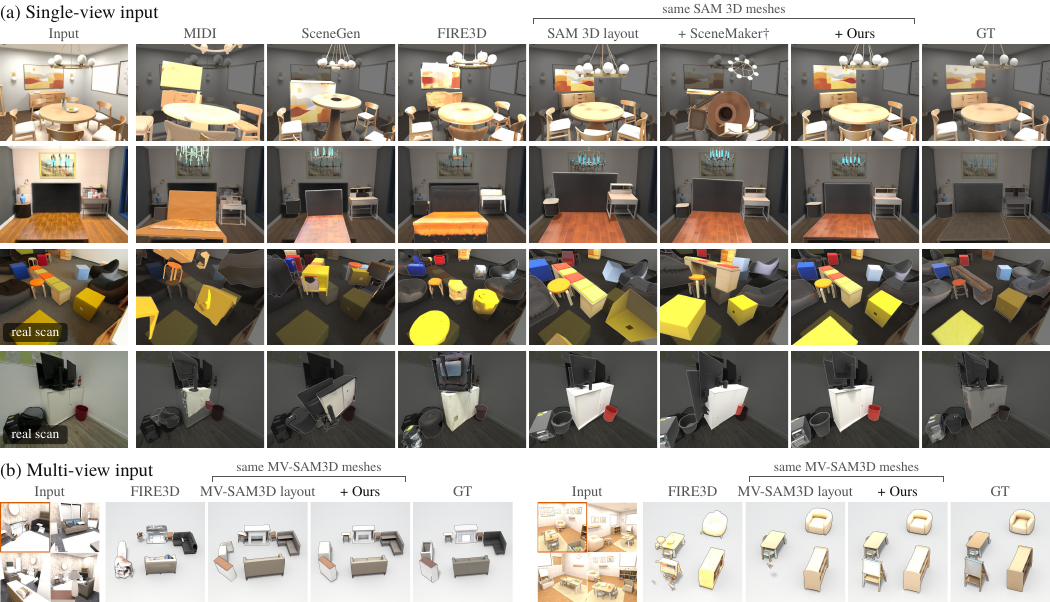}
  \caption{Predicted layouts of generated assets. As in the evaluation (\cref{sec:setup}), each predicted scene is registered to the ground truth before it is shown. In (a), rows 1--2 are synthetic scenes and rows 3--4 are real scans.}
  \label{fig:qual}
\end{figure*}
\paragraph{Training data and sampling.}
\label{sec:data}
Training requires scenes with 9-DoF object annotations. We combine several sources in a common format: synthetic indoor scenes~\cite{fu2021front,fu2021future,khanna2024hssd,zhong2025internscenes,kim2026molmospaces,xia2026sage,pfaff2026scenesmith}, real scans from ScanNet++~\cite{yeshwanth2023scannetpp} with annotated CAD models~\cite{rao2025scannotatepp,chang2015shapenet}, and single-view images~\cite{sun2018pix3d,fu2021future}. We oversample scenes that contain several instances of the same asset, so that the model must rely on the identity embeddings to tell them apart. We also randomly replace the asset of an object with an asset of similar shape from the same category and adjust the labels accordingly, which mimics assets that do not exactly match the object. Since the geometric conditions are optional at inference, each training sample is given a random subset of them. The numbers of scene views, objects, and renders per object are also drawn at random, so one model covers all input settings.

%% file: tables/tab_genasset.tex
\begin{tabular}{llrrrrr<{\hskip\dimexpr\ifdefined\tabgroupgap\tabgroupgap\else 10pt\fi-2\tabcolsep\relax}rrrrr}
\toprule
 &  & \multicolumn{5}{c<{\hskip\dimexpr\ifdefined\tabgroupgap\tabgroupgap\else 10pt\fi-2\tabcolsep\relax}}{\Lmultisource} & \multicolumn{5}{c}{3D-FUTURE} \\
\cmidrule(lr){3-7}\cmidrule(l){8-12}
\Lmeshes & \Llayout & \multicolumn{1}{c}{CD-S$\downarrow$} & \multicolumn{1}{c}{F-S$\uparrow$} & \multicolumn{1}{c}{IoU-B$\uparrow$} & \multicolumn{1}{c}{CD-O$\downarrow$} & \multicolumn{1}{c<{\hskip\dimexpr\ifdefined\tabgroupgap\tabgroupgap\else 10pt\fi-2\tabcolsep\relax}}{F-O$\uparrow$} & \multicolumn{1}{c}{CD-S$\downarrow$} & \multicolumn{1}{c}{F-S$\uparrow$} & \multicolumn{1}{c}{IoU-B$\uparrow$} & \multicolumn{1}{c}{CD-O$\downarrow$} & \multicolumn{1}{c}{F-O$\uparrow$} \\
\midrule
\blockrow{12}{\LsecSV} \\
\addlinespace[1pt]
MIDI~\cite{huang2025midi} & \Lnative & 57.79 & 58.36 & 0.241 & 248.30 & 42.69 & 33.28 & 69.73 & 0.324 & 72.95 & 61.51 \\
 & \Lscenemaker~\cite{shi2025scenemaker} & 12.86 & 84.27 & 0.433 & 18.08 & 79.98 & 11.23 & 88.14 & 0.504 & 22.58 & 85.61 \\
\oursrow  & \Lours & \best{7.29} & \best{92.98} & \best{0.616} & \best{9.06} & \best{92.31} & \best{9.22} & \best{91.10} & \best{0.591} & \best{21.45} & \best{88.66} \\
\addlinespace[2pt]
SceneGen~\cite{meng2026scenegen} & \Lnative & 66.07 & 56.08 & 0.197 & 375.12 & 35.18 & 15.31 & 85.32 & 0.474 & 34.53 & 81.08 \\
 & \Lscenemaker & 13.68 & 83.72 & 0.419 & 20.26 & 79.22 & 10.09 & 89.37 & 0.519 & 22.35 & 87.03 \\
\oursrow  & \Lours & \best{6.49} & \best{92.74} & \best{0.597} & \best{11.40} & \best{91.19} & \best{5.24} & \best{95.58} & \best{0.686} & \best{14.66} & \best{94.45} \\
\addlinespace[2pt]
FIRE3D~\cite{xia2026fire3d} & \Lnative & 11.40 & 89.22 & 0.492 & 19.28 & 87.38 & 13.94 & \best{88.18} & 0.486 & 29.46 & 85.39 \\
 & \Lscenemaker & 15.01 & 82.23 & 0.395 & 22.22 & 77.77 & 15.40 & 83.79 & 0.454 & 28.05 & 81.26 \\
\oursrow  & \Lours & \best{8.95} & \best{90.91} & \best{0.539} & \best{13.45} & \best{89.91} & \best{13.47} & 88.10 & \best{0.509} & \best{24.36} & \best{86.20} \\
\addlinespace[2pt]
SAM 3D~\cite{sam3dteam2025sam3d} & \Lnative & 6.08 & 93.97 & 0.590 & 13.13 & 93.11 & 8.11 & 91.73 & 0.571 & 19.31 & 89.67 \\
 & \Lscenemaker & 12.99 & 86.00 & 0.436 & 22.76 & 81.47 & 10.02 & 89.66 & 0.519 & 21.46 & 87.41 \\
\oursrow  & \Lours & \best{3.43} & \best{97.02} & \best{0.702} & \best{4.86} & \best{96.34} & \best{5.51} & \best{95.14} & \best{0.671} & \best{14.98} & \best{94.03} \\
\midrule
\blockrow{12}{\LsecMV} \\
\addlinespace[1pt]
FIRE3D & \Lnative & 6.73 & 93.59 & 0.572 & 16.56 & 92.30 & -- & -- & -- & -- & -- \\
\oursrow  & \Lours & \best{4.42} & \best{95.29} & \best{0.602} & \best{5.77} & \best{95.28} & -- & -- & -- & -- & -- \\
\addlinespace[2pt]
MV-SAM3D~\cite{li2026mvsam3d} & \Lnative & 6.38 & 94.35 & 0.553 & 11.73 & 93.25 & -- & -- & -- & -- & -- \\
\oursrow  & \Lours & \best{1.61} & \best{98.43} & \best{0.747} & \best{1.49} & \best{98.55} & -- & -- & -- & -- & -- \\
\bottomrule
\end{tabular}

%% file: sec/4_experiments.tex
\section{Experiments}
\label{sec:experiments}

\subsection{Experimental Setup}
\label{sec:setup}

\paragraph{Test data.}
Our main test set, IndoorMix, merges held-out scenes from six indoor datasets: InternScenes~\cite{zhong2025internscenes}, MolmoSpaces~\cite{kim2026molmospaces}, SAGE~\cite{xia2026sage}, SceneSmith~\cite{pfaff2026scenesmith}, 3D-FRONT~\cite{fu2021front}, and the real scans of ScanNet++~\cite{yeshwanth2023scannetpp}. We also report results on the 3D-FUTURE~\cite{fu2021future} test set of SceneGen~\cite{meng2026scenegen}, an existing benchmark of rendered single-view indoor scenes. Details of the test data, results per dataset, and more qualitative results are given in the supplementary material.

\paragraph{Metrics and alignment.}
We use the metrics of MIDI~\cite{huang2025midi}: the scene-level Chamfer distance (CD-S), F-score (F-S), and bounding-box IoU (IoU-B), and, in \cref{tab:genasset}, the object-level CD-O and F-O. The methods compared on generated assets (\cref{tab:genasset}) output layouts in different frames. We follow MIDI and normalize the predicted and the ground-truth scene separately before registering them with a single rigid ICP~\cite{besl1992icp}. These metrics therefore measure the relative layout of the objects. CD-S, a mean over scenes of squared distances, is sensitive to a few scenes with large errors; F-S and IoU-B are less so. With ground-truth CAD models (\cref{tab:cad,tab:pano,tab:ablation}), every method outputs poses in the camera frame of the first scene view, where we compare them without ICP, aligning only the scale of each scene. In all tables, CD values are multiplied by $10^{3}$, F-scores are in percent, and the best result is in bold (\cref{tab:pano,tab:ablation}: second best underlined).\looseness=-1

\begin{table*}[t]
  \centering
  \fontsize{8}{8.4}\selectfont
  \setlength{\tabcolsep}{8pt}\setlength{\tabgroupgap}{27.2pt}\renewcommand{\arraystretch}{1.12}
  \caption{Layout prediction for ground-truth CAD models at their true size. Cond.: given conditions ($K$: intrinsics, $P$: camera poses, $D$: depth). Bold: best per block. For 3D-FUTURE, depth is always estimated.}
  \label{tab:cad}
  \fitbox{\input{tables/tab_cad}}
\end{table*}
\begin{figure*}[t]
  \centering
  \includegraphics[width=1.0\linewidth]{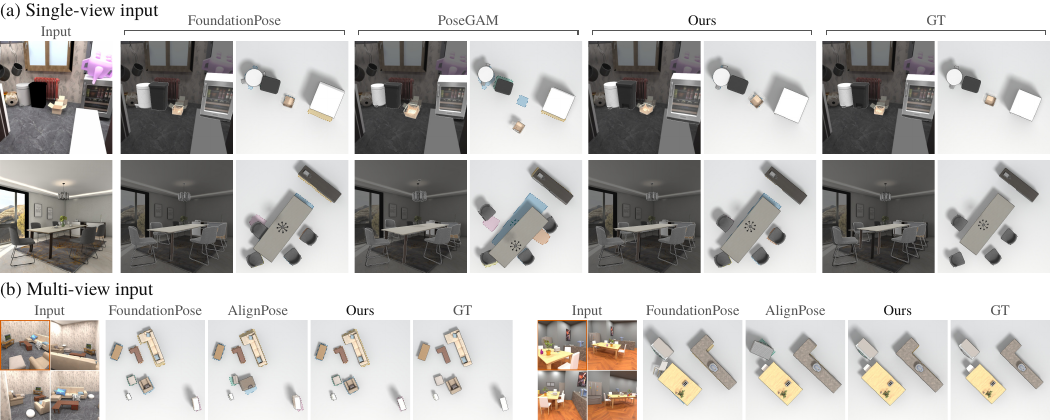}
  \caption{Predicted layouts of ground-truth CAD models. Dashed outlines: ground truth.}
  \label{fig:qualcad}
\end{figure*}
\paragraph{Inputs and implementation.}
All methods that take object masks use the same ground-truth masks. Unless a method specifies its own estimator, estimated geometric conditions come from MoGe-2~\cite{wang2025moge2} for single-view input and from Depth Anything 3~\cite{lin2025depthanything3} for multi-view input. Our model uses the 1B-parameter \vggto{} with rank-32 LoRA and is trained with AdamW~\cite{loshchilov2019adamw} on 8 GPUs for about 200k steps. At inference, we use $M=4$ renders per object.

\subsection{Layout Prediction for Generated Assets}
\label{sec:genasset}

This experiment tests whether a stand-alone layout model places a set of assets more accurately than the method that generated them. We use five scene generation methods: MIDI~\cite{huang2025midi}, SceneGen~\cite{meng2026scenegen}, FIRE3D~\cite{xia2026fire3d}, SAM 3D~\cite{sam3dteam2025sam3d}, and MV-SAM3D~\cite{li2026mvsam3d}. MIDI, SceneGen, and SAM 3D take a single view, MV-SAM3D takes several views, and FIRE3D supports both. Each method generates the meshes of all objects in the scene views, together with its native layout. We then place the same meshes with our model and with the public pose model of SceneMaker~\cite{shi2025scenemaker}, which reports state-of-the-art pose accuracy and takes only a single view.

\paragraph{Results.}
In \cref{tab:genasset}, our layouts are more accurate than SceneMaker's in every single-view group and, with one exception, more accurate than the native layouts in every group and metric: in the FIRE3D group on 3D-FUTURE, the F-S of the native layout is 0.08 points higher. The gain is largest for MIDI and SceneGen, whose native layouts have large errors, but there is also a gain for SAM 3D, whose native layout is already fairly accurate. Averaged over the four single-view groups, our CD-S is 52\% lower than that of SceneMaker on IndoorMix and 28\% lower on 3D-FUTURE. With multi-view input, our layouts improve on the native layouts of FIRE3D and MV-SAM3D in all metrics, most clearly for MV-SAM3D (CD-S 1.61 vs.\ 6.38). Our model receives no geometric conditions in this comparison, whereas FIRE3D, SAM 3D, MV-SAM3D, and SceneMaker all use estimated depth. The gain therefore does not come from a better depth estimator. In \cref{fig:qual}, when objects occlude one another or are cut off by the image border, the baselines place them in the wrong position or orientation, while our layouts stay close to the ground truth.\looseness=-1

\begin{table}[t]
  \centering
  \footnotesize
  \setlength{\tabcolsep}{3pt}
  \caption{Panoramic input: one or two panoramas of a room. For our model, $K$ and $P$ follow from the panorama projection.}
  \label{tab:pano}
  \fitbox{\input{tables/tab_pano_main}}
\end{table}
\subsection{Layout Prediction for GT CAD Models}
\label{sec:cad}

When the asset is the ground-truth CAD model of the object, our task coincides with object pose estimation. We compare with FoundationPose~\cite{wen2024foundationpose}, PoseGAM~\cite{chen2026posegam}, and the multi-view AlignPose~\cite{mikestikova2026alignpose}. Since these methods require true-size CAD models, all methods receive the ground-truth CAD models at their true size, and our model uses the given size in place of its predicted scale.

\paragraph{Results.}
Among the methods that use no ground-truth conditions (upper block of \cref{tab:cad}; examples in \cref{fig:qualcad}), our model without conditions is the most accurate in all three settings. PoseGAM, the only baseline without depth, has much larger errors. With the same estimated conditions as the baselines, our F-S and IoU-B are also higher than those of FoundationPose and AlignPose. Our CD-S is lower than or on par with that of FoundationPose. With $K{+}P{+}D$, however, it is far higher than without conditions (75.95 vs.\ 0.85) and slightly above that of AlignPose. Inaccurate estimated camera poses mislead the model. Ground-truth poses give its lowest CD-S (0.52). Here, estimated conditions do not improve our results and can degrade them, whereas ground-truth ones improve them. With ground-truth conditions (lower block), FoundationPose reaches a higher IoU-B on IndoorMix, as its render-and-compare gains more from accurate depth than our model does. Our F-S under the same conditions is on par or higher.

\begin{figure}[t]
  \centering
  \includegraphics[width=1.0\linewidth]{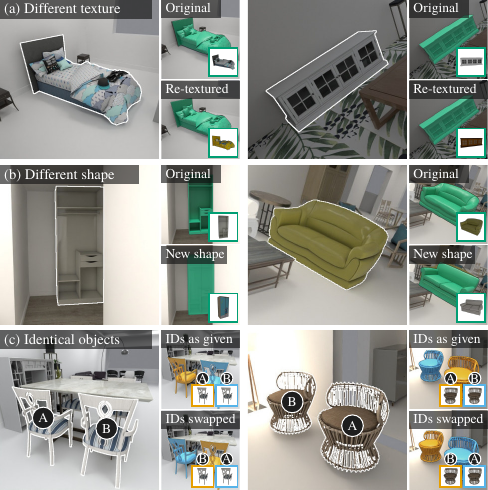}
  \caption{Mismatched assets and identical objects. (a) A re-textured asset is placed like the original one. (b) An asset of a different shape is placed to fit the object. (c) For two identical objects A and B, swapping their identity embeddings swaps the two assets. Large images: input; small: our results (insets: assets).}
  \label{fig:identity}
\end{figure}
\begin{table}[t]
  \centering
  \footnotesize
  \setlength{\tabcolsep}{3pt}
  \caption{Ablation study. Pen.: percentage of objects that penetrate another object by more than 4\,cm.}
  \label{tab:ablation}
  \fitbox{\input{tables/tab_ablation}}
\end{table}
\subsection{Layout Prediction from Panoramic Input}
\label{sec:pano}

We compare with Pano3DComposer~\cite{qiu2026pano3dcomposer} on 42 3D-FRONT houses that are in the training data of neither method, with one or two panoramas of the same room as input. For our model, each panorama is split into six perspective views, whose intrinsics and relative camera poses within the panorama follow from the projection and can be given as conditions. In \cref{tab:pano}, our model is more accurate for both input types, even without conditions. Adding the intrinsics and camera poses derived from a single panorama improves our results further.

\subsection{Ablations and Analysis}
\label{sec:ablation}

\paragraph{Design choices.}
All variants in \cref{tab:ablation} are trained on 3D-FRONT with the same shortened schedule and evaluated there with ground-truth intrinsics and depth. Trained from random initialization, the model fails under this schedule (single-view F-S 40.7 vs.\ 90.8). Without identity embeddings, and hence without the masks, F-S falls to 71.7: the embeddings are needed to tie crops and renders to their objects in general, not only when objects look identical. The object crops give a smaller but consistent gain (single-view CD-S 9.49 vs.\ 11.88). The grounding head helps with multi-view input (CD-S 3.92 vs.\ 4.94) and has a mixed effect with single-view input. Without the physics constraints, the scene metrics are slightly better, but the percentage of penetrating objects rises from 7.9\% to 10.1\%.

\paragraph{Mismatched assets and identical objects.}
Assets from retrieval or generation rarely match the object exactly, and a scene may contain identical objects. We test both cases in \cref{fig:identity}. Re-texturing an asset barely changes the predicted pose (a), and replacing it with another asset of the same category makes the model adapt the pose to the new shape (b). For two identical objects, swapping their identity embeddings swaps the two assets (c), which shows that the model follows the identity embeddings.

\begin{table}[t]
  \centering
  \footnotesize
  \caption{Runtime and peak GPU memory per scene.}
  \label{tab:runtime}
  \input{tables/tab_runtime}
\end{table}
\paragraph{Views, renders, and runtime.}
More scene views improve accuracy at little extra cost. From one to eight views, the time per scene grows by less than 0.2\,s. More renders per object raise F-S and IoU-B slightly (see the supplementary material). In \cref{tab:runtime}, our model takes about 1\,s per scene, and the three baselines take 5.8--7.5\,s.\looseness=-1

%% file: tables/tab_cad.tex
\begin{tabular}{llrrr<{\hskip\dimexpr\ifdefined\tabgroupgap\tabgroupgap\else 8pt\fi-2\tabcolsep\relax}rrr<{\hskip\dimexpr\ifdefined\tabgroupgap\tabgroupgap\else 8pt\fi-2\tabcolsep\relax}rrr}
\toprule
 &  & \multicolumn{3}{c<{\hskip\dimexpr\ifdefined\tabgroupgap\tabgroupgap\else 8pt\fi-2\tabcolsep\relax}}{\Lmultisource, \Lsinglelc} & \multicolumn{3}{c<{\hskip\dimexpr\ifdefined\tabgroupgap\tabgroupgap\else 8pt\fi-2\tabcolsep\relax}}{\Lmultisource, \Lmultilc} & \multicolumn{3}{c}{3D-FUTURE, \Lsinglelc} \\
\cmidrule(lr){3-5}\cmidrule(lr){6-8}\cmidrule(l){9-11}
\Lmethod & \Linput & \multicolumn{1}{c}{CD-S$\downarrow$} & \multicolumn{1}{c}{F-S$\uparrow$} & \multicolumn{1}{c<{\hskip\dimexpr\ifdefined\tabgroupgap\tabgroupgap\else 8pt\fi-2\tabcolsep\relax}}{IoU-B$\uparrow$} & \multicolumn{1}{c}{CD-S$\downarrow$} & \multicolumn{1}{c}{F-S$\uparrow$} & \multicolumn{1}{c<{\hskip\dimexpr\ifdefined\tabgroupgap\tabgroupgap\else 8pt\fi-2\tabcolsep\relax}}{IoU-B$\uparrow$} & \multicolumn{1}{c}{CD-S$\downarrow$} & \multicolumn{1}{c}{F-S$\uparrow$} & \multicolumn{1}{c}{IoU-B$\uparrow$} \\
\midrule
\blockrow{11}{\LsecEst} \\
\addlinespace[1pt]
FoundationPose~\cite{wen2024foundationpose} & $K{+}D$ & 7.51 & 95.19 & 0.739 & 62.96 & 92.96 & 0.758 & 17.35 & 90.73 & 0.654 \\
\addlinespace[2pt]
PoseGAM~\cite{chen2026posegam} & $K$ & 422.10 & 76.77 & 0.549 & 155.98 & 73.34 & 0.445 & 188.58 & 68.12 & 0.452 \\
\addlinespace[2pt]
AlignPose~\cite{mikestikova2026alignpose} & $K{+}P{+}D$ & -- & -- & -- & 71.50 & 91.11 & 0.717 & -- & -- & -- \\
\addlinespace[2pt]
\oursrow \Lours & \Lnone & \best{2.67} & \best{97.42} & \best{0.803} & \best{0.85} & \best{99.42} & \best{0.867} & \best{6.67} & \best{97.53} & \best{0.793} \\
\oursrow  & $K$ & 2.87 & 97.32 & 0.783 & 1.15 & 99.09 & 0.859 & 11.59 & 93.90 & 0.698 \\
\oursrow  & $K{+}D$ & 7.53 & 97.18 & 0.780 & 1.12 & 98.82 & 0.859 & 10.97 & 94.05 & 0.700 \\
\oursrow  & $K{+}P{+}D$ & -- & -- & -- & 75.95 & 98.31 & 0.859 & -- & -- & -- \\
\midrule
\blockrow{11}{\LsecGT} \\
\addlinespace[1pt]
FoundationPose & $K{+}D$ & 2.64 & 98.97 & \best{0.935} & 4.23 & 97.51 & \best{0.905} & 10.62 & 94.58 & 0.759 \\
\addlinespace[2pt]
PoseGAM & $K$ & 439.28 & 76.55 & 0.567 & 100.07 & 81.24 & 0.578 & 147.65 & 72.04 & 0.524 \\
\addlinespace[2pt]
AlignPose & $K{+}P{+}D$ & -- & -- & -- & 8.99 & 96.27 & 0.831 & -- & -- & -- \\
\addlinespace[2pt]
\oursrow \Lours & $K$ & 2.51 & 97.62 & 0.822 & 0.84 & 99.44 & 0.870 & 6.03 & 97.87 & 0.830 \\
\oursrow  & $K{+}D$ & \best{0.95} & \best{99.13} & 0.851 & 0.73 & 99.43 & 0.873 & \best{5.66} & \best{97.92} & \best{0.832} \\
\oursrow  & $K{+}P{+}D$ & -- & -- & -- & \best{0.52} & \best{99.64} & 0.877 & -- & -- & -- \\
\bottomrule
\end{tabular}

%% file: tables/tab_pano_main.tex
\begin{tabular}{llrrr<{\hskip\dimexpr\ifdefined\tabgroupgap\tabgroupgap\else 8pt\fi-2\tabcolsep\relax}rrr}
\toprule
 &  & \multicolumn{3}{c<{\hskip\dimexpr\ifdefined\tabgroupgap\tabgroupgap\else 8pt\fi-2\tabcolsep\relax}}{\Lsinglepano} & \multicolumn{3}{c}{\Lmultipano} \\
\cmidrule(lr){3-5}\cmidrule(l){6-8}
\Lmethod & \Linput & \multicolumn{1}{c}{CD-S$\downarrow$} & \multicolumn{1}{c}{F-S$\uparrow$} & \multicolumn{1}{c<{\hskip\dimexpr\ifdefined\tabgroupgap\tabgroupgap\else 8pt\fi-2\tabcolsep\relax}}{IoU-B$\uparrow$} & \multicolumn{1}{c}{CD-S$\downarrow$} & \multicolumn{1}{c}{F-S$\uparrow$} & \multicolumn{1}{c}{IoU-B$\uparrow$} \\
\midrule
Pano3DComposer~\cite{qiu2026pano3dcomposer} & \Lnone & 25.87 & 84.86 & 0.437 & 14.13 & 89.48 & 0.465 \\
\addlinespace[2pt]
\oursrow \Lours & \Lnone & 5.54 & 96.01 & 0.702 & \second{2.04} & \best{98.69} & \second{0.762} \\
\oursrow  & $K$ & \second{5.13} & \second{96.48} & \second{0.714} & \best{2.01} & \second{98.64} & \best{0.766} \\
\oursrow  & $K{+}P$ & \best{3.83} & \best{96.78} & \best{0.725} & -- & -- & -- \\
\bottomrule
\end{tabular}

%% file: tables/tab_ablation.tex
\begin{tabular}{lrrr<{\hskip\dimexpr\ifdefined\tabgroupgap\tabgroupgap\else 8pt\fi-2\tabcolsep\relax}rrr<{\hskip\dimexpr\ifdefined\tabgroupgap\tabgroupgap\else 8pt\fi-2\tabcolsep\relax}r}
\toprule
 & \multicolumn{3}{c<{\hskip\dimexpr\ifdefined\tabgroupgap\tabgroupgap\else 8pt\fi-2\tabcolsep\relax}}{\Lsingle} & \multicolumn{3}{c<{\hskip\dimexpr\ifdefined\tabgroupgap\tabgroupgap\else 8pt\fi-2\tabcolsep\relax}}{\Lmulti} & \multicolumn{1}{c}{\Lphysics} \\
\cmidrule(lr){2-4}\cmidrule(lr){5-7}\cmidrule(l){8-8}
\Lvariant & \multicolumn{1}{c}{CD-S$\downarrow$} & \multicolumn{1}{c}{F-S$\uparrow$} & \multicolumn{1}{c<{\hskip\dimexpr\ifdefined\tabgroupgap\tabgroupgap\else 8pt\fi-2\tabcolsep\relax}}{IoU-B$\uparrow$} & \multicolumn{1}{c}{CD-S$\downarrow$} & \multicolumn{1}{c}{F-S$\uparrow$} & \multicolumn{1}{c<{\hskip\dimexpr\ifdefined\tabgroupgap\tabgroupgap\else 8pt\fi-2\tabcolsep\relax}}{IoU-B$\uparrow$} & \multicolumn{1}{c}{\Lpen} \\
\midrule
\Lfull & 9.49 & 90.77 & \second{0.578} & \second{3.92} & \second{95.89} & \second{0.579} & \best{7.88} \\
\midrule
\Lnocrops & 11.88 & 89.56 & 0.538 & 5.06 & 95.46 & 0.553 & -- \\
\Lnoid & 72.70 & 71.68 & 0.367 & 58.82 & 72.89 & 0.311 & -- \\
\Lnophys & \best{8.49} & \second{90.92} & \best{0.582} & \best{3.67} & \best{96.65} & \best{0.586} & 10.10 \\
\Lnoground & \second{8.86} & \best{90.98} & 0.568 & 4.94 & 95.51 & 0.555 & -- \\
\Lnopre & 128.84 & 40.74 & 0.160 & 275.97 & 35.46 & 0.073 & -- \\
\bottomrule
\end{tabular}

%% file: tables/tab_runtime.tex
\begin{tabular}{lrr}
\toprule
\Lmethod & \multicolumn{1}{c}{\Ltime$\downarrow$} & \multicolumn{1}{c}{\Lmem$\downarrow$} \\
\midrule
FoundationPose~\cite{wen2024foundationpose} & 7.5 & \best{4.3} \\
PoseGAM~\cite{chen2026posegam} & 6.8 & 15.5 \\
SceneMaker~\cite{shi2025scenemaker} & 5.8 & 13.6 \\
\addlinespace[2pt]
\oursrow \Lours & \best{0.91} & 5.2 \\
\bottomrule
\end{tabular}

%% file: sec/5_conclusion.tex
\section{Conclusion}
\label{sec:conclusion}

We presented X-3DComposer, a feed-forward model that composes a 3D scene from given assets by predicting their layout. Its key idea is to treat objects as views: with asset renders as extra views, layout becomes camera prediction, a task that pretrained multi-view geometry models already solve well. Our experiments show that layout need not be tied to asset generation. On the same meshes, our layouts are more accurate than the native layouts of five scene generation methods. Without depth input, our scene-level Chamfer distance is 52\% lower than that of SceneMaker on IndoorMix and 28\% lower on 3D-FUTURE. As the model needs only asset renders, it can be combined with any generator or retrieval system. Our model has limitations. It requires object masks as input and is trained and tested only on indoor scenes. It is also sensitive to inaccurate geometric conditions, and its physics constraints reduce penetration without guaranteeing physically plausible layouts. Predicting the masks within the model and extending it to other scene types are left for future work.

%% file: sec/X_suppl.tex
\clearpage
\setcounter{page}{1}
\maketitlesupplementary
\appendix
\renewcommand{\dblfloatpagefraction}{0.7}

Section and equation numbers, Tables 1--5, and Figures 1--6 refer to the main paper, and the tables follow its conventions.

\section{Implementation Details}
\label{sec:supp_impl}

\paragraph{Backbone.}
The backbone is the 1B-parameter \vggto{} model. It has a token width of 1024 and alternates 24 frame-attention blocks with 24 global-attention blocks; five of the global-attention blocks attend only to the camera and register tokens. As in the original model, the resolution of a scene view depends on its aspect ratio, and every scene view has about $416^2$ pixels (for example, $416\times416$ or $304\times560$). For frame attention, views are batched by resolution.

\paragraph{New modules.}
The ray encoder is a convolution of stride 16 on the 6-channel Pl\"ucker rays, followed by a linear layer that maps to the token width. For a view with intrinsics but no camera pose, two learnable ``camera pose unknown'' vectors are added, one to its camera token and the other to each of its patch tokens. The depth encoder is a second convolution of stride 16, applied to the logarithm of the normalized depth and a validity mask. The camera encoder is an MLP that maps the field of view and the world-to-camera rotation (as a quaternion) and translation to the token width. There are 64 identity embeddings of dimension 64, so one forward pass handles at most 64 objects. In addition to the patch-wise identity vectors of \cref{eq:identity}, the camera token of each crop and render receives the embedding of its object. The output layers of the ray and camera encoders, the two ``camera pose unknown'' vectors, the linear projections of the identity vectors, and the three log-scale outputs added to the camera head are zero-initialized. The convolution of the depth encoder has no bias and starts from small random weights; it adds nothing to the tokens of views without depth input.

\paragraph{Frames and geometric conditions.}
In the asset frame, $+Y$ points up, and coordinates have no physical unit. The scale $\bs_i$ and the translation $\bt_i$ share one unit of length, which can be fixed only by depth or by camera poses with a nonzero baseline. Camera poses and depth are accepted only together with intrinsics, so there are five combinations of conditions: $K{+}P{+}D$, $K{+}P$, $K{+}D$, $K$, and none.

\paragraph{Normalization and metric-scale recovery.}
Let $\mathcal{X}$ be the ground-truth scene points (in the frame of the first camera) and $\bo_j$ the camera center of scene view $j$ ($\bo_1=\mathbf{0}$). The training targets are divided by $g$, and, following OmniVGGT~\cite{peng2025omnivggt}, the input depth and camera centers are normalized separately:
\begin{equation}
\begin{gathered}
  g=\frac{1}{|\mathcal{X}|}\sum_{\bx\in\mathcal{X}}\lVert\bx\rVert_2,\qquad
  \tilde D=\frac{D}{\operatorname{mean}(D)},\\
  \tilde\bo_j=\frac{\bo_j}{\beta},\qquad \beta=\frac{1}{S-1}\sum_{j=2}^{S}\lVert\bo_j\rVert_2 ,
\end{gathered}
\end{equation}
with $\beta=1$ if $S=1$ or if all camera centers coincide. At inference, the global factor of \cref{sec:oav}, denoted here by $\lambda$, is $\lambda_D$ if depth is given, and otherwise $\lambda_P$ if camera poses of several views are given:
\begin{equation}
  \lambda_D=\operatorname{median}\big(D/\hat D\big),\qquad
  \lambda_P=\frac{\sum_{j\ge2}\lVert\bo_j\rVert_2\,\lVert\hat\bo_j\rVert_2}{\sum_{j\ge2}\lVert\hat\bo_j\rVert_2^2},
\end{equation}
where the median is taken over the pixels with valid input depth, and $\hat D$ and $\hat\bo_j$ are the predicted depth and camera centers.

\paragraph{Training objective.}
The total loss is
\begin{equation}
\begin{aligned}
  \mathcal{L}={}&\mathcal{L}_{\mathrm{cam}}+5\mathcal{L}_{s}+15\mathcal{L}_{\mathrm{shape}}+\mathcal{L}_{\mathrm{cons}}\\
  &+2\mathcal{L}_{\mathrm{cand}}+\mathcal{L}_{\mathrm{fuse}}+\mathcal{L}_{\mathrm{agree}}+0.5\,\mathcal{L}_{\mathrm{crop}}\\
  &+0.5\,\mathcal{L}_{\mathrm{gr}}+0.3\,\mathcal{L}_{\mathrm{floor}}+\mathcal{L}_{\mathrm{pen}} ,
\end{aligned}
  \label{eq:supp_loss}
\end{equation}
where $\mathcal{L}_{\mathrm{cam}}$ contains the original losses of the backbone on the scene views and the camera loss of the renders (\cref{sec:training}), and $\mathcal{L}_{\mathrm{crop}}$ is the sum of the camera and depth losses of the crops. The camera label of a crop is a virtual camera with a centered principal point, fitted by minimizing the reprojection error of the visible surface points of the object in the crop. With $\overline{\,\cdot\,}$ the mean over the three axes, $\Pi(\mathbf{v})=\mathbf{v}-\overline{\mathbf{v}}\,\mathbf{1}$, the three scale terms are
\begin{equation}
\begin{aligned}
  \mathcal{L}_{s}&=\mathbb{E}^{w}_{i,m}\,\overline{\lvert\bl_{i,m}-\log\bs_i^*\rvert},\\
  \mathcal{L}_{\mathrm{shape}}&=\mathbb{E}^{w}_{i,m}\,\overline{\lvert\Pi(\bl_{i,m})-\Pi(\log\bs_i^*)\rvert},\\
  \mathcal{L}_{\mathrm{cons}}&=\mathbb{E}_{i,m}\,\overline{(\bl_{i,m}-\log\bs_i)^2},
\end{aligned}
\end{equation}
and the three pose terms are
\begin{equation}
\begin{aligned}
  \mathcal{L}_{\mathrm{cand}}&=\mathbb{E}_{i,m}\,d_i\big(R_{i,m},\bt_{i,m};\,R_i^*,\bt_i^*\big),\\
  \mathcal{L}_{\mathrm{fuse}}&=\mathbb{E}_{i}\,d_i\big(R_i,\bt_i;\,R_i^*,\bt_i^*\big),\\
  \mathcal{L}_{\mathrm{agree}}&=\mathbb{E}_{i,m}\,d_i\big(R_{i,m},\bt_{i,m};\,R_i,\bt_i\big).
\end{aligned}
\end{equation}
Here $\mathbb{E}^{w}$ is a mean weighted by $w_i\propto1+2\,\mathrm{clip}\big(1-a_i/0.01,\,0,\,1\big)$, where $a_i$ is the largest fraction of a scene view covered by the visible mask of object $i$; a small object thus gets up to three times the weight of the others. The distance $d_i$ is defined in \cref{eq:pose_err}, and $\bs_i$ is the predicted scale (\cref{sec:oav}), which $\mathcal{L}_{\mathrm{fuse}}$ and $\mathcal{L}_{\mathrm{agree}}$ use in \cref{eq:vcam}. For $\mathcal{L}_{\mathrm{cand}}$, the scale in \cref{eq:vcam} is $\gamma_i\bs_i^*$ with $\gamma_i=\exp\overline{(\log\bs_i-\log\bs_i^*)}$, so these candidates use the ground-truth axis ratios and the geometric mean of the predicted scale.

\emph{Symmetry.} We detect the symmetry group $G_i$ of each asset about its vertical axis from points sampled on its surface (a discrete $n$-fold symmetry with $n\ge2$, or a continuous rotational symmetry). During training, the ground-truth rotation is replaced by $R_i^*Q^\star$ with $Q^\star=\arg\min_{Q\in G_i}\sum_m\lVert R_{i,m}-R_i^*Q\rVert_F$, and the camera labels of the renders change accordingly.

\emph{Grounding head.} For a crop or render $r$ of object $i$, let $\mathbf{c}_r$ be its final camera token and $\mathbf{h}_u$ the final token of scene-view patch $u$. Then
\begin{equation}
\begin{gathered}
  p_r(u)=\operatorname*{softmax}_u\big(\tau\cos(W_1\mathbf{c}_r,\,W_2\mathbf{h}_u)\big),\\
  q_r(u)=\frac{\alpha_{i,u}}{\sum_{u'}\alpha_{i,u'}},\qquad
  \mathcal{L}_{\mathrm{gr}}=\mathbb{E}_r\,\mathrm{KL}\big(q_r\,\Vert\,p_r\big),
\end{gathered}
\end{equation}
where $W_1$ and $W_2$ are linear maps, $\tau$ is a learnable temperature, and $\alpha_{i,u}$ is as in \cref{eq:identity}.

\emph{Physics.} Both terms are computed from the fused poses, with lengths in meters. For a pair of objects $(i,k)$, we sample points $\mathbf{y}\in\mathcal{Y}_i$ inside object $i$, each standing for a volume $v_{\mathbf{y}}$, and read their penetration depth $\delta_k(\mathbf{y})=[-\phi_k(\mathbf{y})]_+$ from the precomputed signed distance field $\phi_k$ of object $k$. The floor term covers the set $\mathcal{B}$ of objects that are floor-supported in the ground truth; $b_i$ is the height of the bottom of object $i$, and $e_i$ is its offset from the floor, defined below:
\begin{equation}
\begin{aligned}
  \mathcal{L}_{\mathrm{pen}}&=\frac{w_{\mathrm{src}}}{N}\sum_{i\ne k}\,\sum_{\mathbf{y}\in\mathcal{Y}_i}v_{\mathbf{y}}\big[\delta_k(\mathbf{y})-\epsilon_{ik}\big]_+,\\
  \mathcal{L}_{\mathrm{floor}}&=\mathbb{E}_{i\in\mathcal{B}}\big[\lvert e_i\rvert-1\,\mathrm{cm}\big]_+ .
\end{aligned}
\end{equation}
The tolerance is $\epsilon_{ik}=\max(1.5\,\mathrm{cm},\,\delta^*_{ik}+1\,\mathrm{cm})$, where $\delta^*_{ik}$ is the largest penetration depth of the same pair in the ground-truth layout. $w_{\mathrm{src}}$ is 0.06 for 3D-FRONT samples, 0.009 for 3D-FUTURE samples, and 0 for the other sources. If the ground truth of the sample gives the floor height $b_0$, $e_i=b_i-b_0$; otherwise $e_i=(b_i-b_i^*)-\operatorname{median}_{k\in\mathcal{B}}(b_k-b_k^*)$.

\paragraph{Training samples.}
We describe the sampling of the final model (\cref{sec:supp_generic}); the differences for the main-paper model are given at the end. Each sample is given intrinsics with probability 0.8. When intrinsics are given, camera poses and depth are each given independently with probability 0.5, so each of the five combinations of conditions has probability 0.2. At each step, we draw the number of scene views $S$ (1 to 10) and of objects $N$ (1 to 12) from a fixed set of combinations, and the number of renders per object $M$ from $\{1,2,3,4,6,8\}$. Each object gets one crop from each scene view in which the long side of its input mask is at least 64 pixels; if there are more than three such views, the three with the longest sides are kept. A crop whose box side is at least 336 pixels is used at $336\times336$, the others at $256\times256$; the renders of an object with a 336-pixel crop are also 336 pixels, and the other renders are 256 pixels. The batch size is set by a budget of about 56k tokens per GPU, and all samples of one step have the same number of views at each resolution. In the multi-view sources, any two scene views $a$ and $b$ of a sample satisfy
\begin{equation}
  \lVert\bo_a-\bo_b\rVert_2\ \ge\ \max\big(0.6\,\mathrm{m},\ 0.3\,\bar z_{ab}\big),
\end{equation}
where $\bar z_{ab}$ is the mean over the two views of the median depth of the visible objects; pairs whose camera centers are less than 0.2\,m apart but whose optical axes differ by at least $30^\circ$ are exempt. A target object must be visible in at least one of these views, with a visible area of at least 0.25\% of the image and with at most 55\% of the object occluded. Renders are composited onto a uniform gray background, and their appearance is randomly perturbed. At inference, the crops are chosen from the input masks by the same rule, and the renders use fixed viewpoints spread evenly in azimuth around the asset, at elevations alternating between $10^\circ$ and $25^\circ$. The main-paper model differs mainly in: one or two 256-pixel crops per object (one at inference), one to five renders, and about 28k tokens per GPU and step.

\paragraph{Optimization.}
We use AdamW ($\beta_1=0.9$, $\beta_2=0.95$, weight decay 0.05) with gradient clipping at a norm of 1. The peak learning rate is $2.5\times10^{-5}$ for the new modules and $5\times10^{-6}$ for the LoRA weights (rank 32) and the output branch of the camera head. The learning rate is warmed up linearly for 500 steps and then decays with a cosine schedule to 1\% of its peak. Training uses 8 GPUs and runs for about 200k steps.

\section{Training Data}
\label{sec:supp_data}

All data are converted to one format. Each scene stores the images, object masks, and cameras of its views and, where available, their depth, and each asset is rendered in advance from 32 viewpoints in its asset frame. We render the 3D-FRONT rooms with Blender from several viewpoints along a walkable trajectory in each room and take the labels from the layout files. The other synthetic scenes come from HSSD, InternScenes, MolmoSpaces, SAGE, and SceneSmith; we convert their labels from the object transformations that each source provides. For ScanNet++, we use the DSLR images together with the ShapeNet~\cite{chang2015shapenet} CAD models and the pose annotations of SCANnotate++~\cite{rao2025scannotatepp}. These CAD models differ in proportions from the real objects, and we take the per-axis scale of each object from the annotations. The single-view images are real photos from Pix3D and rendered scenes from 3D-FUTURE. For the asset replacement of \cref{sec:training}, the substitute is the asset of the same category with the most similar shape; during training, an object that has a substitute is replaced by it with probability 0.5. \Cref{tab:traindata} lists the sources.

\begin{table}[t]
  \centering
  \footnotesize
  \setlength{\tabcolsep}{4pt}
  \caption{Training sources. Scenes: number of training scenes (images for the single-view sources). Share: share of the training samples. The share of 3D-FRONT includes 9.2 percentage points of samples drawn to contain several instances of the same asset.}
  \label{tab:traindata}
  \begin{tabular}{@{}llrr@{}}
    \toprule
    Source & Type & Scenes & Share \\
    \midrule
    3D-FRONT & synthetic, multi-view & 17,814 & 42.3\% \\
    InternScenes & synthetic (from real scans), multi-view & 6,030 & 9.2\% \\
    SAGE & synthetic, multi-view & 1,908 & 5.5\% \\
    MolmoSpaces & synthetic, multi-view & 1,238 & 6.4\% \\
    HSSD & synthetic, multi-view & 919 & 6.4\% \\
    SceneSmith & synthetic, multi-view & 290 & 3.7\% \\
    ScanNet++ & real, multi-view & 222 & 8.0\% \\
    3D-FUTURE & synthetic, single-view & 13,776 & 11.0\% \\
    Pix3D & real, single-view & 5,829 & 7.4\% \\
    \bottomrule
  \end{tabular}
\end{table}

\section{Evaluation Details}
\label{sec:supp_eval}
\begin{table}[t]
  \centering
  \scriptsize
  \setlength{\tabcolsep}{2.5pt}
  \caption{Results of \cref{tab:genasset} by dataset (F-S and IoU-B). $\dagger$: pose model only. Bold: best in each row.}
  \label{tab:persource}
  \fitbox{\input{tables/tab_persource}}
\end{table}
\begin{table*}[t]
  \centering
  \footnotesize
  \setlength{\tabcolsep}{4pt}\setlength{\tabgroupgap}{10pt}
  \caption{Our model without conditions and with estimated $K{+}D$ (Cond.: given conditions), in the setting of \cref{tab:genasset}. Bold: better of the two; values that are equal after rounding are not bolded.}
  \label{tab:genasset_est}
  \fitbox{\input{tables/tab_genasset_est}}
\end{table*}
\begin{table}[t]
  \centering
  \footnotesize
  \setlength{\tabcolsep}{3pt}
  \caption{Effect of the number of scene views $S$ and of renders per object $M$ (81 scenes, no conditions). Time: seconds per scene on a single GPU.}
  \label{tab:views}
  \fitbox{\input{tables/tab_views}}
\end{table}

\begin{figure*}[p]
  \centering
  \includegraphics[width=\linewidth,height=0.86\textheight,keepaspectratio]{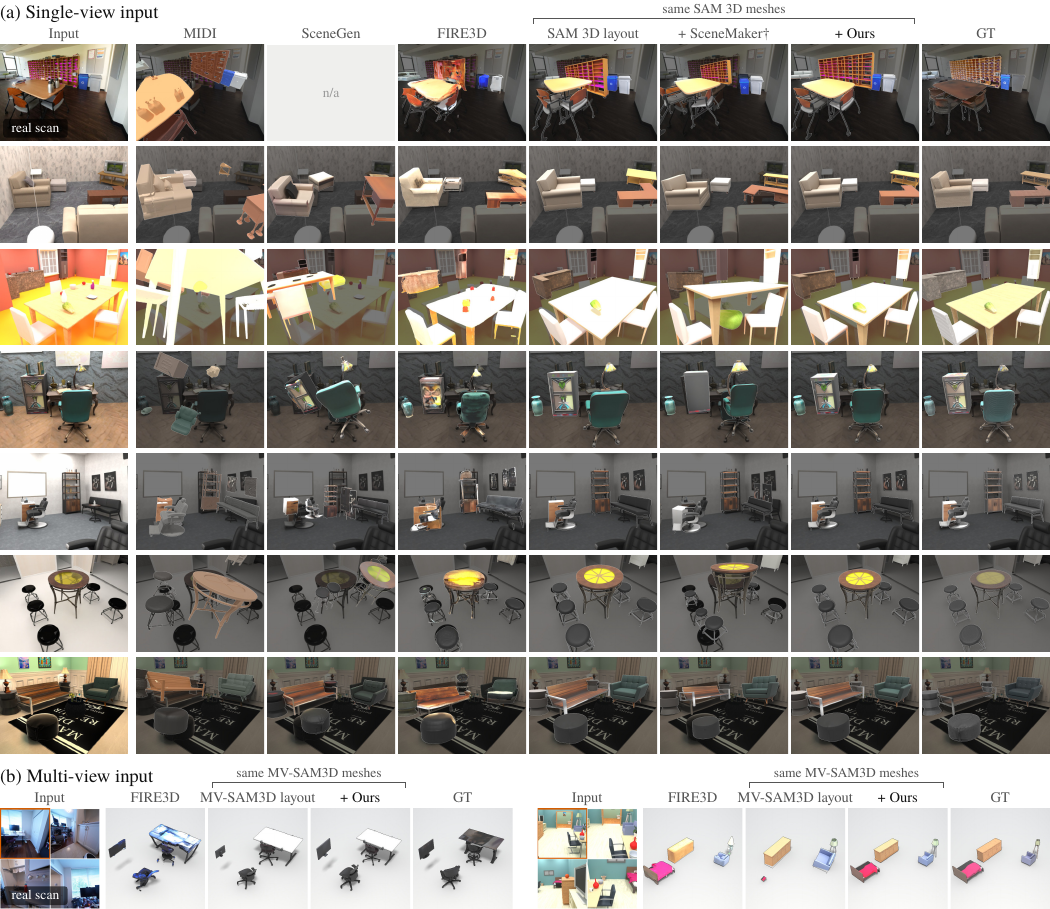}
  \caption{Predicted layouts of generated assets: further cases in the setting of \cref{fig:qual}. Single-view rows, from top to bottom: a real scan from ScanNet++, InternScenes, MolmoSpaces, SAGE, SceneSmith, 3D-FRONT, and 3D-FUTURE. Multi-view: a real scan from ScanNet++ and MolmoSpaces. In the real scans, every visible object is placed; for objects without a CAD annotation, the ground truth shows the corresponding segment of the scanned mesh. n/a: SceneGen produced no output for this scene.}
  \label{fig:qual_supp}
\end{figure*}
\begin{figure*}[t]
  \centering
  \includegraphics[width=\linewidth]{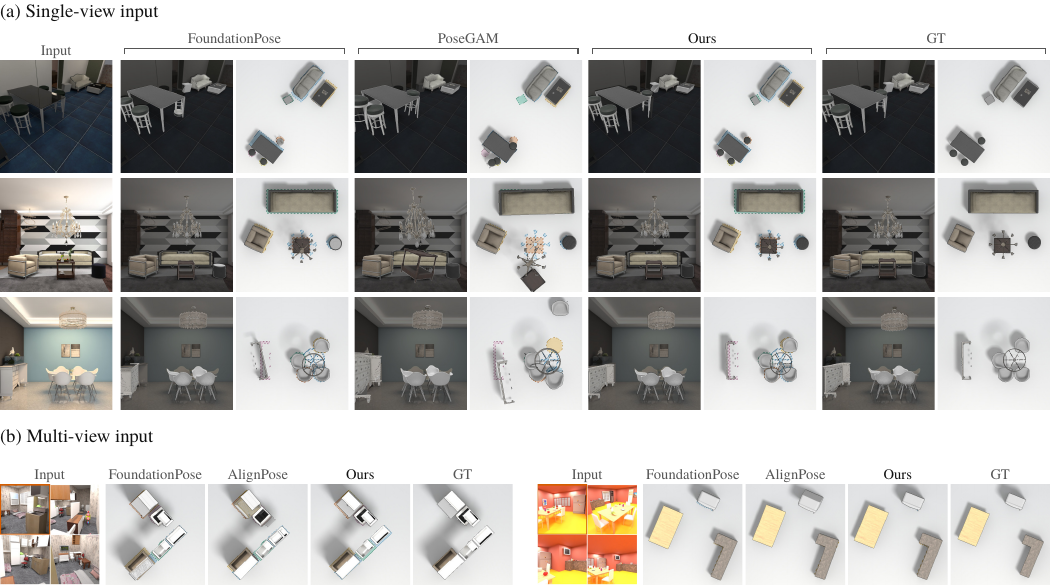}
  \caption{Predicted layouts of ground-truth CAD models with estimated conditions. (a) Single-view input: one scene from 3D-FRONT and two from 3D-FUTURE, each method shown in the camera view and in a top view. (b) Multi-view input (top views): one scene from InternScenes and one from MolmoSpaces. Dashed outlines: ground truth.}
  \label{fig:qualcad_supp}
\end{figure*}
\begin{figure*}[t]
  \centering
  \includegraphics[width=\linewidth]{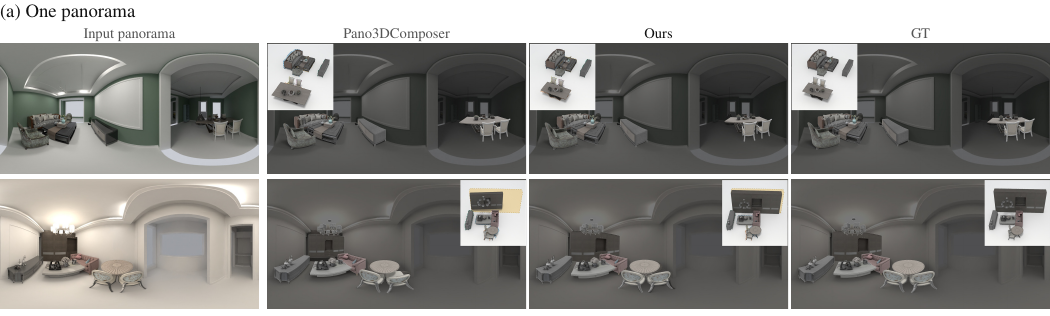}\\[4pt]
  \includegraphics[width=\linewidth]{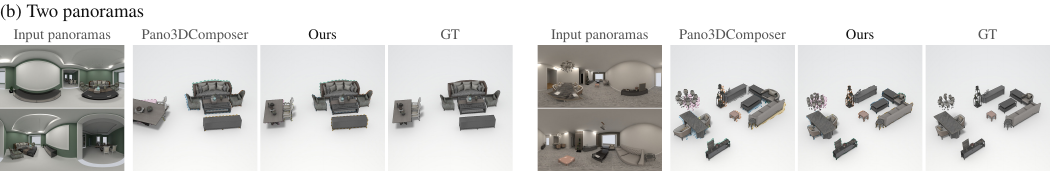}
  \caption{Panoramic input. (a) One panorama: objects overlaid on the panorama, with an oblique view as inset (dashed: ground truth); our model uses $K{+}P$. (b) Two panoramas of the same room: oblique views; our model uses $K$.}
  \label{fig:pano}
\end{figure*}
\begin{figure*}[t]
  \centering
  \includegraphics[width=\linewidth]{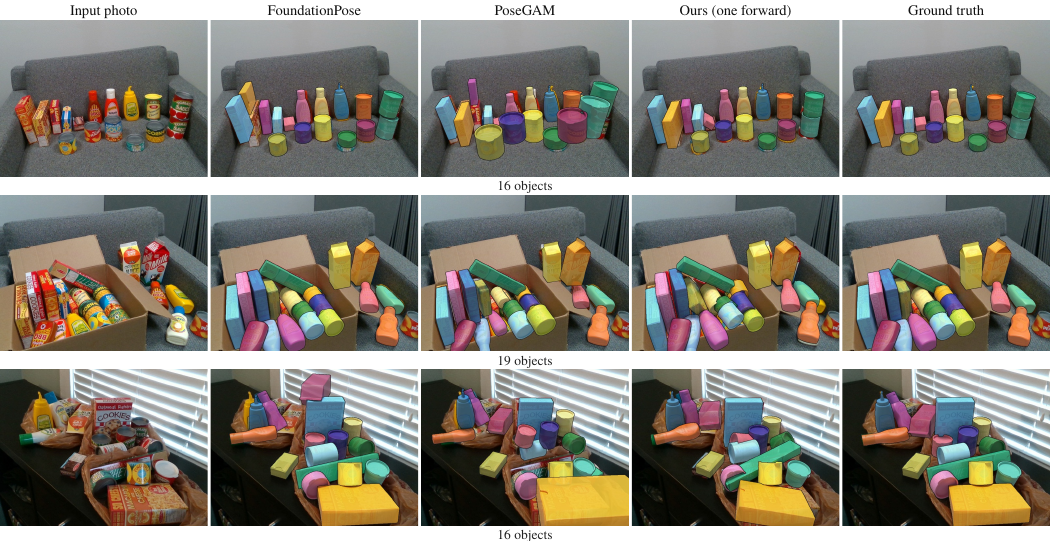}
  \caption{Comparison with FoundationPose and PoseGAM on real HOPE photos. All methods use the same photo, intrinsics, ground-truth masks, and true-size object models; FoundationPose and our final model also use the sensor depth. FoundationPose and PoseGAM estimate the poses object by object; our model places all objects of a photo in one forward pass, with the given sizes. The assets are overlaid on the photo in their own colors, and each object has the same color in all columns.}
  \label{fig:generic_real}
\end{figure*}
\begin{figure*}[t]
  \centering
  \includegraphics[width=\linewidth]{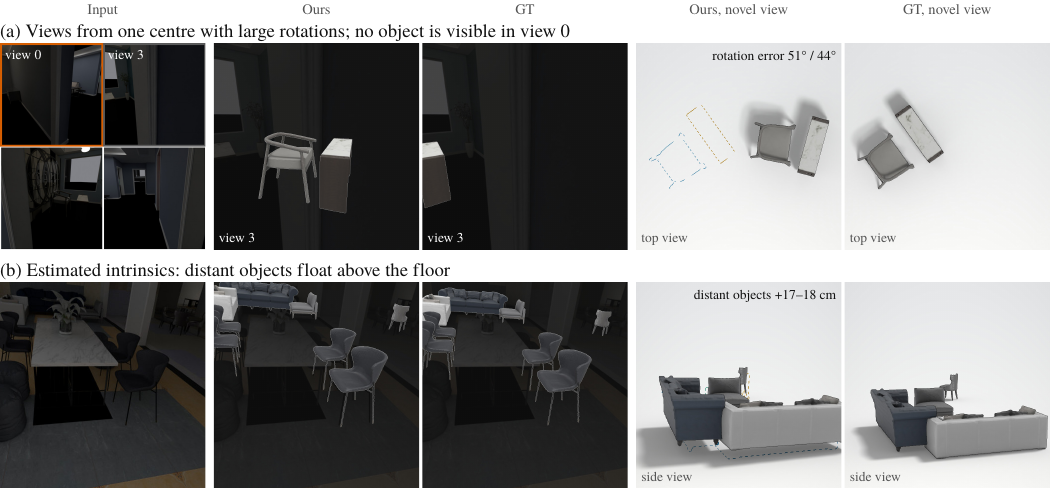}
  \caption{Failure cases. Columns: input, our result, ground truth, and novel views of the latter two (dashed: ground-truth objects). (a) No object is visible in the first view, and the views share one camera center. (b) Inaccurate estimated intrinsics.}
  \label{fig:failure}
\end{figure*}

\paragraph{Test data.}
IndoorMix consists of 400 single-view scenes and 224 multi-view scenes with 2, 4, 8, or 12 views. On the scenes evaluated in \cref{tab:genasset}, the single-view scenes of IndoorMix have 4.5 target objects on average and the multi-view scenes 4.8, with at most 8 in both; the scenes of 3D-FUTURE have 3.6. About 12\% of the IndoorMix scenes are real scans from ScanNet++, and a further 11\% of the single-view and 20\% of the multi-view scenes come from the part of InternScenes that is reconstructed from real scans. The MolmoSpaces scenes come from its ProcTHOR part. Multi-view scenes must have co-visible views: we connect two views if at least 1\% of the ground-truth depth points of one are visible and unoccluded in the other, or if both show the same target object with at least 100 pixels each, and we keep a scene only if its view graph is connected, which removes 10 of 234 candidate scenes. This rule only selects test scenes and does not change the input of any method. Each comparison is computed on the scenes for which every compared method returns the poses of all objects; \cref{tab:persource} gives the number of scenes per dataset.

\paragraph{Overlap with training data.}
The IndoorMix scenes are held out from training, but all six datasets are among our training sources, whereas most baselines were trained on at most one of them. MIDI and SceneMaker were trained on 3D-FRONT, so some of our 3D-FRONT test rooms may be in their training sets. The training data of FIRE3D include SAGE, InternScenes, and SceneSmith scenes and 3D-FUTURE assets and therefore also overlap with our test datasets. The 3D-FUTURE test set uses the same CAD models as 3D-FRONT, and some of them also appear in the 3D-FUTURE training images, on which both SceneGen and our model are trained.

\paragraph{Metrics and alignment.}
As in MIDI, each scene is centered and scaled so that the longest side of its bounding box is 1.9, and all metrics are computed in this frame. CD is a squared distance, and the F-score threshold of 0.1 is about 5\% of this longest side. CD-S and F-S are computed on the whole scene and averaged over scenes. IoU-B is the IoU between the axis-aligned 3D bounding boxes of each predicted object and its ground-truth object, and CD-O and F-O are the Chamfer distance and the F-score of each object; these three metrics are measured after the scene-level alignment, without any per-object re-alignment, and are averaged over all objects of the test set. For generated meshes, all methods are evaluated with the same code, which follows the official evaluation of MIDI; we correct a transposition error in the Kabsch step of its ICP.

With ground-truth CAD models, the prediction and the ground truth are both normalized with the transformation of the ground-truth scene, and only the scale is aligned. Each predicted scene gets one evaluation scale
\begin{equation}
  \kappa=\operatorname*{median}_{i}\frac{\lVert\bt_i^*\rVert_2}{\lVert\bt_i\rVert_2},
\end{equation}
the median ratio between the ground-truth and the predicted distance from the first camera to the object centers. In \cref{tab:cad}, the predicted object centers are replaced by $\kappa\,\bt_i$ and the objects keep their size; in \cref{tab:pano,tab:ablation,tab:views}, the whole predicted layout is scaled by $\kappa$ about the first camera. Translation and rotation are not aligned, so global translation and rotation errors of the layout count toward the metrics. The evaluation scale is unrelated to the global factor $\lambda$ of \cref{sec:oav}.

\paragraph{Inputs of the compared methods.}
MoGe-2~\cite{wang2025moge2} estimates intrinsics and depth and is not given the ground-truth field of view. Depth Anything 3~\cite{lin2025depthanything3} estimates intrinsics, depth, and camera poses relative to the first scene view. In \cref{tab:genasset}, MIDI and SceneGen use only images and masks, and SAM 3D and SceneMaker use point maps from MoGe~\cite{wang2025moge}. MV-SAM3D uses the estimates of Depth Anything 3. FIRE3D uses MoGe-2 for single-view input and, for multi-view input, the same Depth Anything 3 estimates as MV-SAM3D.

\paragraph{Ground-truth CAD models.}
For our model, the given size replaces the predicted scale only in the final layout; the rotation and translation are still those that the model predicts. In the rows without depth input, the translations are multiplied by the median ratio between the given sizes and the predicted scales, which sets the unit of length of the scene. FoundationPose is a single-view method. On multi-view scenes, it estimates the pose of each object in the view where the object's mask is largest, and this pose is transferred to the first scene view with the camera pose. We scale the thresholds of its two depth pre-filters with the distance of the object, because the fixed thresholds of the official code can remove all depth pixels of a distant object, in which case no pose is estimated. AlignPose starts from multi-view FoundationPose results and refines the pose of each object jointly over the views, using the given intrinsics and camera poses (those of Depth Anything 3 in the upper block of \cref{tab:cad}); it is therefore listed with $K{+}P{+}D$. PoseGAM uses no depth and handles multi-view scenes in the same way as FoundationPose. For FoundationPose and PoseGAM, the camera pose is used only for this change of frame and is not listed as a condition. We correct an error in the translation reset of the official PoseGAM code, which combines the intrinsics of the image crop with mask coordinates in the full image.

\paragraph{Panoramic input.}
The test data are 42 houses of 3D-FRONT: the 12 houses of the Pano3DComposer test split that are not in our training data, and 30 further held-out houses that are in neither the training nor the test split of Pano3DComposer. With one panorama, each capture position is a sample; with two panoramas, a sample consists of two capture positions in the same room. The six perspective views of a panorama share its center; they are $60^\circ$ apart in yaw, are pitched $15^\circ$ downward, and have a field of view of $72^\circ$. Their intrinsics and relative camera poses follow exactly from this arrangement; since the views share one camera center, these poses fix no unit of length. The relative camera pose between two panoramas is unknown, and our model accepts camera poses only for all views or for none, so for two panoramas we report only the settings without conditions and with intrinsics. Two panoramas give 12 scene views, more than the maximum of 10 used in training. Among the six views of the first panorama, the one in which the most objects are visible is used as the first scene view; this choice depends only on the input masks. Pano3DComposer estimates depth and the relative camera poses of the panoramas with its built-in Depth Anything 3.

\section{Additional Results}
\label{sec:supp_results}

\paragraph{Results by dataset.}
The results of \cref{tab:genasset} are broken down by dataset in \cref{tab:persource}. Our layouts are the best in both metrics in 37 of the 40 rows. In the other three rows, all in the two FIRE3D groups, one metric is slightly lower than that of the native layout. Our layouts are the best in every group on the real scans of ScanNet++ and also on 3D-FRONT, the training set of MIDI and SceneMaker.

\paragraph{Layout prediction for generated assets with estimated conditions.}
\Cref{tab:genasset_est} compares our model without conditions and with estimated $K{+}D$ in the setting of \cref{tab:genasset}. The two settings differ little in the scene-level metrics; with estimated $K{+}D$, CD-O is lower in most groups.

\paragraph{Number of views and renders.}
\Cref{tab:views} reports results for different numbers of scene views $S$ and of renders per object $M$ on 81 multi-view scenes from InternScenes, MolmoSpaces, SAGE, SceneSmith, and ScanNet++, with ground-truth CAD models and without conditions. For each scene, we use its first $S$ views and the same objects for all $S$. More views give more accurate layouts. More renders raise F-S and IoU-B, with a small gain from four to eight renders, while the time grows from 0.54\,s to 0.90\,s. CD-S does not decrease monotonically with $M$; it is sensitive to a few scenes with large errors.

\paragraph{More qualitative results.}
\Cref{fig:qual_supp,fig:qualcad_supp,fig:pano} show further cases with generated assets, with ground-truth CAD models and estimated conditions, and with panoramic input, in the settings of \cref{fig:qual}, \cref{tab:cad}, and \cref{tab:pano}.

\section{Extension to Generic Objects}
\label{sec:supp_generic}

The main-paper model is trained and tested on indoor scenes, where most assets are furniture. To test whether the approach carries over to other kinds of objects, we continue training the main-paper model with generic-object data added to the training set: the physically based rendered (PBR) training images of BOP~\cite{hodan2018bop} (T-LESS~\cite{hodan2017tless}, HomebrewedDB~\cite{kaskman2019homebreweddb}, HOPE~\cite{tyree2022hope}, HANDAL~\cite{guo2023handal}, YCB-V~\cite{xiang2018posecnn}, and LINEMOD~\cite{hinterstoisser2012linemod}), tabletop scenes that we generate by physics simulation with objects from GSO~\cite{downs2022gso} and ShapeNet~\cite{chang2015shapenet}, and real images from GraspNet-1Billion~\cite{fang2020graspnet} and YCB-V. In the last part of this training, these data make up 19\% of the training samples (11.5\% BOP renderings, 6.0\% tabletop scenes, and 1.5\% real images). The architecture and the training objective are unchanged, and the samples are drawn as described in \cref{sec:supp_impl}. We call the result the final model; it is used only in this section. As in the main paper, the inputs include ground-truth object masks.

\paragraph{Comparison on real photos.}
\Cref{fig:generic_real} compares the final model with FoundationPose~\cite{wen2024foundationpose} and PoseGAM~\cite{chen2026posegam} on real photos from the HOPE validation set. The photos are not in the training set, but their objects were seen in training. Each photo contains 12 to 21 objects, whereas a training sample contains at most 12. All three methods are given the same inputs (listed in the caption); as in \cref{tab:cad}, our model uses the given sizes in place of its predicted scales, and the rotations and translations are its own predictions. On the photos shown, our results are visually similar to those of FoundationPose, which renders and refines each object iteratively, whereas PoseGAM, which uses no depth, errs more along the viewing direction. Unlike FoundationPose, our model uses no test-time optimization, and unlike both methods, it gives the poses of all objects in one forward pass.

\section{Limitations and Failure Cases}
\label{sec:supp_qual}
\label{sec:limitations}

All our experiments use ground-truth object masks; the effect of predicted masks remains untested. The main-paper model is trained and tested only on indoor scenes; generic objects are covered in \cref{sec:supp_generic}.

\emph{Orientation of cameras and objects.} The cameras of the indoor training data are never rolled far about the optical axis. In our tests, rolling the camera by $180^\circ$ lowers the accuracy sharply, and objects lying on their side or upside down are placed less accurately.

\emph{Reference frame.} As in the backbone, all outputs are expressed in the camera frame of the first scene view, so the result depends on which view comes first. In \cref{fig:failure}(a), the six views come from one panorama: they share one camera center, differ by large rotations, and no object is visible in the first view; the predicted object rotations have large errors. This is why the first view of the panorama experiments is chosen from the input masks (\cref{sec:supp_eval}).

\emph{Geometric conditions.} The model is sensitive to inaccurate geometric conditions (\cref{tab:cad}). In \cref{fig:failure}(b), the intrinsics estimated by MoGe-2 are inaccurate, and distant objects float above the floor.

\emph{Small objects.} A scene view has about $416^2$ pixels, so small distant objects cover few pixels; the crops compensate only in part.

\emph{Physical plausibility.} The physics constraints reduce penetration between objects but do not guarantee physically plausible layouts.

%% file: tables/tab_persource.tex
\begin{tabular}{lllrrrrrr}
\toprule
 &  &  & \multicolumn{2}{c}{\Lnative} & \multicolumn{2}{c}{\Lscenemaker~\cite{shi2025scenemaker}} & \multicolumn{2}{c}{\Lours} \\
\cmidrule(lr){4-5}\cmidrule(lr){6-7}\cmidrule(l){8-9}
\Lmeshes & \Ldataset & \Lscenes & \multicolumn{1}{c}{F-S$\uparrow$} & \multicolumn{1}{c}{IoU-B$\uparrow$} & \multicolumn{1}{c}{F-S$\uparrow$} & \multicolumn{1}{c}{IoU-B$\uparrow$} & \multicolumn{1}{c}{F-S$\uparrow$} & \multicolumn{1}{c}{IoU-B$\uparrow$} \\
\midrule
\blockrow{9}{\LsecSV} \\
\addlinespace[1pt]
MIDI~\cite{huang2025midi} & \Lreconscans & 36 & 57.9 & 0.232 & 82.0 & 0.398 & \best{89.6} & \best{0.553} \\
 & MolmoSpaces & 37 & 41.6 & 0.120 & 84.5 & 0.306 & \best{94.9} & \best{0.561} \\
 & SAGE & 38 & 58.8 & 0.227 & 89.8 & 0.448 & \best{94.8} & \best{0.599} \\
 & SceneSmith & 39 & 67.1 & 0.293 & 85.6 & 0.414 & \best{95.9} & \best{0.658} \\
 & ScanNet++ & 39 & 55.3 & 0.248 & 75.5 & 0.416 & \best{86.1} & \best{0.566} \\
 & 3D-FRONT & 146 & 61.1 & 0.263 & 85.3 & 0.481 & \best{93.9} & \best{0.651} \\
 & 3D-FUTURE & 134 & 69.7 & 0.324 & 88.1 & 0.504 & \best{91.1} & \best{0.591} \\
\addlinespace[2pt]
SceneGen~\cite{meng2026scenegen} & \Lreconscans & 36 & 54.6 & 0.156 & 83.2 & 0.370 & \best{90.1} & \best{0.507} \\
 & MolmoSpaces & 36 & 37.8 & 0.090 & 83.6 & 0.311 & \best{92.8} & \best{0.493} \\
 & SAGE & 38 & 56.8 & 0.177 & 89.1 & 0.431 & \best{96.2} & \best{0.611} \\
 & SceneSmith & 39 & 58.2 & 0.211 & 83.3 & 0.411 & \best{92.7} & \best{0.587} \\
 & ScanNet++ & 36 & 49.0 & 0.197 & 75.9 & 0.425 & \best{85.4} & \best{0.561} \\
 & 3D-FRONT & 146 & 61.9 & 0.237 & 84.5 & 0.457 & \best{94.3} & \best{0.652} \\
 & 3D-FUTURE & 134 & 85.3 & 0.474 & 89.4 & 0.519 & \best{95.6} & \best{0.686} \\
\addlinespace[2pt]
FIRE3D~\cite{xia2026fire3d} & \Lreconscans & 36 & 88.9 & 0.472 & 82.8 & 0.381 & \best{89.7} & \best{0.500} \\
 & MolmoSpaces & 37 & 92.4 & 0.420 & 81.5 & 0.283 & \best{94.5} & \best{0.483} \\
 & SAGE & 38 & 90.5 & 0.492 & 85.8 & 0.401 & \best{92.7} & \best{0.542} \\
 & SceneSmith & 38 & 94.0 & 0.539 & 83.6 & 0.384 & \best{94.7} & \best{0.567} \\
 & ScanNet++ & 38 & 84.1 & 0.503 & 76.6 & 0.415 & \best{86.2} & \best{0.548} \\
 & 3D-FRONT & 141 & 88.2 & 0.502 & 82.5 & 0.427 & \best{90.1} & \best{0.555} \\
 & 3D-FUTURE & 131 & \best{88.2} & 0.486 & 83.8 & 0.454 & 88.1 & \best{0.509} \\
\addlinespace[2pt]
SAM 3D~\cite{sam3dteam2025sam3d} & \Lreconscans & 36 & 94.4 & 0.573 & 84.3 & 0.396 & \best{95.2} & \best{0.643} \\
 & MolmoSpaces & 37 & 97.3 & 0.536 & 85.3 & 0.323 & \best{99.1} & \best{0.684} \\
 & SAGE & 38 & 96.1 & 0.596 & 91.3 & 0.451 & \best{98.3} & \best{0.684} \\
 & SceneSmith & 39 & 97.5 & 0.663 & 85.8 & 0.421 & \best{98.8} & \best{0.779} \\
 & ScanNet++ & 39 & 89.1 & 0.551 & 79.5 & 0.438 & \best{93.5} & \best{0.655} \\
 & 3D-FRONT & 146 & 92.8 & 0.597 & 87.0 & 0.476 & \best{97.1} & \best{0.717} \\
 & 3D-FUTURE & 134 & 91.7 & 0.571 & 89.7 & 0.519 & \best{95.1} & \best{0.671} \\
\midrule
\blockrow{9}{\LsecMV} \\
\addlinespace[1pt]
FIRE3D & \Lreconscans & 43 & 95.2 & 0.611 & -- & -- & \best{96.0} & \best{0.638} \\
 & MolmoSpaces & 43 & 94.7 & 0.502 & -- & -- & \best{95.2} & \best{0.510} \\
 & SAGE & 39 & 95.7 & \best{0.636} & -- & -- & \best{96.7} & 0.630 \\
 & SceneSmith & 37 & \best{96.5} & 0.685 & -- & -- & 96.3 & \best{0.697} \\
 & ScanNet++ & 26 & 85.5 & 0.459 & -- & -- & \best{91.0} & \best{0.551} \\
 & 3D-FRONT & 26 & 89.9 & 0.500 & -- & -- & \best{94.9} & \best{0.587} \\
\addlinespace[2pt]
MV-SAM3D~\cite{li2026mvsam3d} & \Lreconscans & 41 & 93.7 & 0.554 & -- & -- & \best{97.6} & \best{0.757} \\
 & MolmoSpaces & 41 & 96.6 & 0.543 & -- & -- & \best{99.0} & \best{0.682} \\
 & SAGE & 35 & 95.0 & 0.590 & -- & -- & \best{99.0} & \best{0.779} \\
 & SceneSmith & 35 & 95.2 & 0.646 & -- & -- & \best{99.6} & \best{0.818} \\
 & ScanNet++ & 26 & 90.4 & 0.456 & -- & -- & \best{96.2} & \best{0.686} \\
 & 3D-FRONT & 26 & 93.6 & 0.504 & -- & -- & \best{98.9} & \best{0.764} \\
\bottomrule
\end{tabular}

%% file: tables/tab_genasset_est.tex
\begin{tabular}{llrrrrr<{\hskip\dimexpr 10pt-2\tabcolsep\relax}rrrrr}
\toprule
 &  & \multicolumn{5}{c<{\hskip\dimexpr 10pt-2\tabcolsep\relax}}{\Lmultisource} & \multicolumn{5}{c}{3D-FUTURE} \\
\cmidrule(lr){3-7}\cmidrule(l){8-12}
\Lmeshes & \Linput & \multicolumn{1}{c}{CD-S$\downarrow$} & \multicolumn{1}{c}{F-S$\uparrow$} & \multicolumn{1}{c}{IoU-B$\uparrow$} & \multicolumn{1}{c}{CD-O$\downarrow$} & \multicolumn{1}{c<{\hskip\dimexpr 10pt-2\tabcolsep\relax}}{F-O$\uparrow$} & \multicolumn{1}{c}{CD-S$\downarrow$} & \multicolumn{1}{c}{F-S$\uparrow$} & \multicolumn{1}{c}{IoU-B$\uparrow$} & \multicolumn{1}{c}{CD-O$\downarrow$} & \multicolumn{1}{c}{F-O$\uparrow$} \\
\midrule
\blockrow{12}{\LsecSV} \\
\addlinespace[1pt]
MIDI~\cite{huang2025midi} & \Lnone & 7.29 & 92.98 & 0.616 & 9.06 & 92.31 & 9.22 & 91.10 & 0.591 & 21.45 & 88.66 \\
 & $K{+}D$ & \best{7.14} & \best{93.05} & \best{0.620} & \best{8.40} & \best{92.53} & \best{8.71} & \best{91.84} & \best{0.598} & \best{20.55} & \best{89.69} \\
\addlinespace[2pt]
SceneGen~\cite{meng2026scenegen} & \Lnone & \best{6.49} & 92.74 & 0.597 & 11.40 & 91.19 & \best{5.24} & 95.58 & \best{0.686} & 14.66 & 94.45 \\
 & $K{+}D$ & 6.71 & \best{92.95} & \best{0.601} & \best{10.84} & \best{91.55} & 5.27 & \best{95.75} & 0.681 & \best{14.00} & \best{94.55} \\
\addlinespace[2pt]
FIRE3D~\cite{xia2026fire3d} & \Lnone & 8.95 & \best{90.91} & 0.539 & 13.45 & \best{89.91} & \best{13.47} & 88.10 & 0.509 & \best{24.36} & 86.20 \\
 & $K{+}D$ & 8.95 & 90.54 & \best{0.544} & \best{10.94} & 89.77 & 13.73 & \best{88.47} & \best{0.513} & 24.69 & \best{86.54} \\
\addlinespace[2pt]
SAM 3D~\cite{sam3dteam2025sam3d} & \Lnone & 3.43 & 97.02 & 0.702 & 4.86 & 96.34 & 5.51 & 95.14 & 0.671 & 14.98 & 94.03 \\
 & $K{+}D$ & \best{3.17} & \best{97.15} & 0.702 & \best{3.86} & \best{96.56} & \best{5.03} & \best{95.88} & \best{0.675} & \best{14.35} & \best{94.83} \\
\midrule
\blockrow{12}{\LsecMV} \\
\addlinespace[1pt]
FIRE3D & \Lnone & 4.42 & 95.29 & 0.602 & 5.77 & 95.28 & -- & -- & -- & -- & -- \\
 & $K{+}D$ & \best{4.40} & \best{95.38} & \best{0.608} & \best{5.66} & \best{95.38} & -- & -- & -- & -- & -- \\
\addlinespace[2pt]
MV-SAM3D~\cite{li2026mvsam3d} & \Lnone & 1.61 & 98.43 & \best{0.747} & \best{1.49} & 98.55 & -- & -- & -- & -- & -- \\
 & $K{+}D$ & 1.61 & \best{98.44} & 0.742 & 1.51 & \best{98.57} & -- & -- & -- & -- & -- \\
\bottomrule
\end{tabular}

%% file: tables/tab_views.tex
\begin{tabular}{rrrrr<{\hskip\dimexpr 12pt-2\tabcolsep\relax}rrrrr}
\toprule
\multicolumn{5}{c<{\hskip\dimexpr 12pt-2\tabcolsep\relax}}{(a) $S$ ($M=4$)} & \multicolumn{5}{c}{(b) $M$ ($S=4$)} \\
\cmidrule(r){1-5}\cmidrule(l){6-10}
\multicolumn{1}{c}{$S$} & \multicolumn{1}{c}{CD-S$\downarrow$} & \multicolumn{1}{c}{F-S$\uparrow$} & \multicolumn{1}{c}{IoU-B$\uparrow$} & \multicolumn{1}{c<{\hskip\dimexpr 12pt-2\tabcolsep\relax}}{\Ltime} & \multicolumn{1}{c}{$M$} & \multicolumn{1}{c}{CD-S$\downarrow$} & \multicolumn{1}{c}{F-S$\uparrow$} & \multicolumn{1}{c}{IoU-B$\uparrow$} & \multicolumn{1}{c}{\Ltime} \\
\midrule
1 & 4.93 & 96.25 & 0.789 & 0.47 & 1 & 2.63 & 97.90 & 0.835 & 0.25 \\
2 & 2.79 & 98.32 & 0.847 & 0.50 & 2 & \second{1.39} & 98.97 & 0.867 & 0.34 \\
4 & \second{1.58} & \second{99.01} & \second{0.873} & 0.54 & 4 & 1.58 & \second{99.01} & \second{0.873} & 0.54 \\
8 & \best{0.95} & \best{99.44} & \best{0.886} & 0.65 & 8 & \best{1.10} & \best{99.28} & \best{0.884} & 0.90 \\
\bottomrule
\end{tabular}